\documentclass[]{fairmeta}

\usepackage{amsmath}
\usepackage{amssymb}
\usepackage{graphicx}
\usepackage{booktabs}
\usepackage{dsfont}
\usepackage[most]{tcolorbox}
\usepackage{multirow}
\usepackage{xspace}
\usepackage{tabularx}
\tcbuselibrary{listings, breakable}

\usepackage{algorithm}
\usepackage{algorithmic}

\usepackage{enumitem}
\usepackage{float}
\usepackage{xcolor}
\usepackage{tcolorbox}

\definecolor{findingnavy}{RGB}{0,0,128}
\newtcolorbox{findingbox}{
  colback=gray!10,
  colframe=gray!10,
  boxrule=0pt,
  arc=3pt,
  left=8pt, right=8pt, top=3pt, bottom=3pt,
  before skip=6pt,
  after skip=6pt,
  before upper={\textcolor{findingnavy}{\bfseries Finding:}\ },
}

\makeatletter
\renewcommand\paragraph{\@startsection{paragraph}{4}{\z@}%
  {1.5ex \@plus 0.5ex \@minus .2ex}%
  {-1em}%
  {\normalfont\normalsize\bfseries}}
\makeatother

\makeatletter
\newcommand{\printappendixtoc}{%
  \begingroup
  \par\medskip
  \begin{tcolorbox}[
    colback=metabg,
    colframe=metablue,
    arc=2mm,
    boxrule=0.5pt,
    left=6pt,right=6pt,top=4pt,bottom=4pt
  ]
    {\sffamily\bfseries Appendix Contents}\par\smallskip
    {\small\@starttoc{atoc}}%
  \end{tcolorbox}
  \endgroup
}
\makeatother

\newtcblisting{promptbox}{%
  colback=gray!4,%
  colframe=gray!45,%
  arc=1mm,%
  boxrule=0.5pt,%
  left=5pt,right=5pt,top=4pt,bottom=4pt,%
  enhanced,%
  breakable,%
  listing only,%
  listing options={%
    basicstyle=\scriptsize\ttfamily,%
    columns=fullflexible,%
    keepspaces=true,%
    breaklines=true,%
    breakatwhitespace=false,%
    breakindent=0pt,%
    postbreak=\mbox{\textcolor{gray}{$\hookrightarrow$}\space},%
    upquote=true,%
    literate={—}{{-{}-{}-}}3 {–}{{-{}-}}2 {≠}{{!=}}2 {⚠}{{!}}1%
  }%
}

\newcommand{\promptlabel}[1]{\par\medskip\noindent\textbf{#1}\par\nopagebreak\smallskip}

\title{Thinking Before Thinking: Scaling Agentic Inference Through Meta-Reasoning}

\author[1]{Paras Dahal}
\author[1]{Anton Bakhtin}
\author[1]{Taco Cohen}
\author[1]{Zhengxing Chen}
\author[1]{Carole-Jean Wu}
\author[1]{Rob Fergus}
\author[1]{Scott Yih}
\author[1]{Gabriel Synnaeve}
\author[1]{Ruslan Salakhutdinov}
\author[1]{Sanjeev Arora}
\author[1]{Jason Weston}
\author[1]{Anirudh Goyal}

\affiliation[1]{Meta Superintelligence Labs}

\date{\today}
\correspondence{padahal@meta.com, agi@meta.com}

\abstract{
As agents take on longer and more complex problems, controlling the execution becomes a task in its own right. Each step in the run brings new control choices, like which partial work to build on, whether to start fresh, or when to stop. We introduce \emph{agentic meta-reasoning}, an inference-time harness that makes these choices an explicit and structured reasoning process. Workers carry out the task-level computation, while a controller consolidates what the run has established, explores next options, assesses what each option is worth under the remaining budget, and dispatches the chosen work with context drawn from persistent memory. Between decisions the controller carries only a compact account of the run rather than replaying its full history. Our baselines span production coding agents and research harnesses, together with a Direct Control Agent using the same workers and compute budget allowance. On ProgramBench, which tests long-horizon agentic capability through program reconstruction, meta-reasoning achieves 71.5\% with GPT-5.5 against 58.0\% for Codex; with Opus 4.8 it achieves 67.2\% against 65.5\% for Claude Code. On the other benchmarks, spanning abstract reasoning, multi-domain long-horizon reasoning, and proof generation, it gains between 3.6 and 4.2 points over direct control, averaged across three frontier models. It keeps improving over the tested budget ranges where direct control plateaus, though its overhead can hurt at small budgets. Artifact-graph analysis reveals more reuse of earlier work, higher coverage of correct solutions in most settings, and nonuniform gains in final selection. These results indicate that spending computation on structured control becomes more important as agents scale to longer runs.
}

\begin{document}
\maketitle

\begin{figure}[!htbp]
    \centering
    \includegraphics[width=\linewidth]{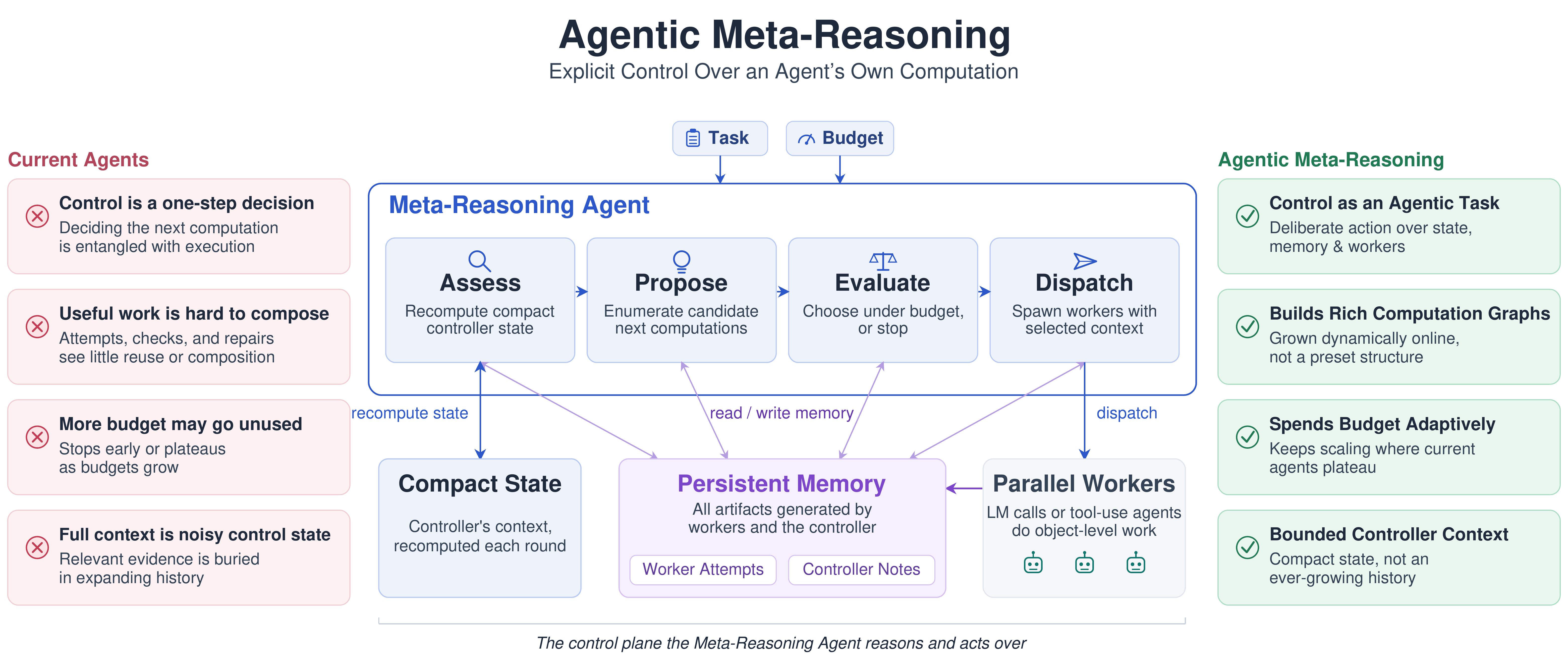}
    \caption{
    \textbf{Overview of agentic meta-reasoning.} Current agent designs interleave two cognitive functions, control and object-level work. This makes useful work hard to compose, and performance hard to scale as more compute becomes available. Agentic meta-reasoning makes control its own explicit, structured reasoning process over compact state, persistent memory, and workers.
    }
    \label{fig:framework}
\end{figure}

\section{Introduction}
\label{sec:introduction}

Language model agents increasingly spend a large number of model calls on a single problem, and a considerable share of those calls are decisions about what to do next. Consider a math agent working on a proof, holding a candidate whose main argument looks sound but whose central lemma is unverified. It can attempt the lemma, ask a second worker to check the argument it already has, or abandon the approach for a different one. Each option costs more computation, and those spent on the wrong choice may be wasted. For an agent working autonomously toward a goal, this choice is part of solving the problem, and a capability distinct from producing the next step of the proof.

This is a problem of \emph{metacognitive control}: assessing one's own progress and using that assessment to decide what to do next~\citep{desabbata2024rationalmetareasoning,liu2026metacognitionsurvey}. In a single model call, the object of that control is the chain of thought. In an agent it is the work the run has already produced, so it must decide which results to trust, what to build on, and when to stop~\citep{li2025meco,xiang2026dtsr}. A mistaken judgment costs more than the compute it consumes. It can keep a failed approach alive for the rest of the run, or throw away a correct answer the agent has already found~\citep{brown2024monkeys}. Existing agents often interleave these decisions with the object-level work, taking each control decision in a single step conditioned on their accumulating history~\citep{yao2023react,shinn2023reflexion,shen2023hugginggpt,packer2023memgpt,zhang2025recursive}. In this paper we argue that these decisions deserve an agentic reasoning process of their own. An agent can deliberate and inquire before it commits, reading back an earlier result or asking a worker to check one. That deliberation has a structure of its own, consolidating what the run has established, exploring what could be done next, and assessing what each option is worth.

We call this approach \emph{agentic meta-reasoning}: the agent reasons about and acts on its own inference process. We implement it as a new inference-time \emph{harness}, the runtime that coordinates model calls and manages the run. Our Meta-Reasoning Agent separates the workers, which do the task-level work, from the controller, which decides what work to assign. The controller keeps a compact assessment of progress. Full worker outputs remain in persistent memory, where they can be retrieved when needed.

Each control cycle runs four stages: the controller updates the state based on what has changed, proposes next computations, evaluates their value under the remaining budget, and dispatches the chosen work along with the earlier outputs each worker should see. Each of these stages is a full agentic process, with its own instructions and tools. The controller model calls are charged against the same budget as the workers', so the extra deliberation has to earn what it costs. We record the resulting work as an \emph{artifact graph}. Every stored output is an artifact, and an edge marks one artifact being supplied as context for producing another. The graph shows where a run branches and which later steps build on earlier ones, letting us diagnose it by the work it produced rather than by its final score alone.

We evaluate this framework on IMO ProofBench-Advanced, ARC-AGI-2, LongCoT-mini, and ProgramBench, spanning proof search, abstract reasoning, long-horizon reasoning, and program reconstruction, using Gemini 3.1 Pro, GPT-5.5, and Opus 4.8. A worker is either a single model call or a coding agent that uses tools. Our baselines span production coding agents and research harnesses, together with a Direct Control Agent that holds workers and actions fixed so that only the control differs, choosing each action in a single call over accumulated history.

At the main allowances of 100 model calls on the reasoning benchmarks and 1200 on ProgramBench, meta-reasoning yields higher point estimates in all 12 matched comparisons, with mean gains of 3.6--4.2 score points by benchmark. On ProgramBench with GPT-5.5, mean per-problem test-pass rate rises to 71.5\% as compared to 58.0\% for Codex and 63.7\% for direct control. The performance keeps improving as the allowance grows where direct control plateaus, and the artifact graphs show more reuse of earlier work and a correct candidate present in more runs. These results indicate that agentic meta-reasoning, with explicit separation of control and object-level work, leads to agents that scale better as the task length and complexity grow.

\section{Related Work}
\label{sec:related}

\paragraph{Test-time computation and agentic harnesses.}
Inference-time methods allocate computation along a single reasoning path, across independent samples followed by selection, through repeated critique and revision, or over a predefined search structure~\citep{wei2022chain,wang2022selfconsistency,madaan2023selfrefine,yao2023tree,besta2024graph}. These established inference-time computation as a major scaling lever, but they fix the structure of the computation, the number of samples, branches, refinement rounds, verifier calls, or aggregation steps, before the problem reveals what kind of work it needs. Agentic harnesses instead use the model's own judgments to direct next computations, interleaving reasoning with environment actions, feeding back on earlier attempts, delegating subtasks to specialist models, or decomposing a task into a plan before execution~\citep{yao2023react,shinn2023reflexion,shen2023hugginggpt,prasad2024adapt,kim2024llmcompiler}, and modern coding agents extend this to long-running software tasks with file inspection, editing, and command execution~\citep{anthropic2026claudecode,openai2025codex}. However, as a task produces more intermediate artifacts, the context these agents act from grows noisier~\citep{bertsch2025oolongevaluatinglongcontext}, which in turn affects their control decisions. This motivates our view of agentic inference as online artifact-graph construction, where the controller reads the graph to direct work and workers extend it.

\paragraph{Learned and optimized orchestration.}
A complementary line of work learns, searches, or optimizes the system that coordinates model calls. GPTSwarm learns communication structures among agents~\citep{zhuge2024gptswarm}; AgentOrchestra and related systems study multi-agent coordination~\citep{zhang2025agentorchestra}; Meta-Harness and Fugu optimize or train harnesses and orchestrator models for stronger task performance~\citep{lee2026metaharness,sakana2026fugu}; Conductor and Trinity similarly treat orchestration as a first-class object of system design~\citep{nielsen2025conductor,xu2025trinity}. More recent work automates harness construction directly, synthesizing, evolving, or learning to edit it from execution feedback and failure trajectories~\citep{lou2026autoharnessimprovingllmagents,lin2026agenticharnessengineeringobservabilitydriven,shao2026harnessr1learningeditexecutable}; \citet{ren2026selfimprovementsmodernagenticsystems} survey the area. Whether such automation improves on simpler test-time scaling is under active investigation~\citep{wang2026rethinkingevaluationharnessevolution}. Rather than training a new orchestrator or searching over a task-specific harness, we isolate the control mechanism and structure it as its own agentic process.

\paragraph{Metacognitive control of inference.}
The metacognition literature gives a more precise vocabulary for this control problem: an intelligent system must monitor the state of its own computation and use that monitoring to regulate what it does next~\citep{liu2026metacognitionsurvey}. Work operationalizing this for language models casts inference as a budgeted value-of-computation problem, turns self-monitoring signals such as feeling-of-knowing or judgment-of-learning into trust and retry decisions, separates object-level reasoning from meta-level regulation, adds an explicit monitoring stage to generate-verify pipelines, and uses metacognitive signals to decide when to invoke tools or stop reasoning~\citep{desabbata2024rationalmetareasoning,zhao2026roireasoning,cao2026knowact,dong2025metar1,oh2025mgv,li2025meco,xiang2026dtsr}. Our setting differs in the object being controlled: not a single chain, tool trigger, or stopping rule, but a growing external computation over persistent artifacts.

\paragraph{Memory and context as control.}
Long-running agents also require mechanisms for preserving and selecting information. MemGPT treats memory management as part of the language-model system~\citep{packer2023memgpt}; A-MEM, Mem0, and ACE develop adaptive or persistent memory interfaces for agents~\citep{xu2025amem,chhikara2025mem0,zhang2025ace}; CoALA frames agents as systems with structured memory, action, and decision components~\citep{sumers2024coala}. The Recursive Language Model (RLM) holds the context in a variable the model edits with code rather than in a transcript~\citep{zhang2025recursive}. PRO-LONG keeps a complete interaction log and searches it programmatically~\citep{fox2026prolongprogrammaticmemoryenables}. In our setting memory is an action surface for controlling inference, since the controller writes artifacts, reads selected prior work, and chooses what context each worker sees, determining what later computation can build on.

\paragraph{Diagnosing agent runs.}
Several recent works move beyond final accuracy by asking where performance comes from. Large Language Monkeys separates generating a correct candidate from selecting it~\citep{brown2024monkeys}; Other work studies the structure of generated candidate sets and search traces~\citep{saadfalcon2025weaver,dragoi2025beyondpassk}; AgentEval, MAST, and Who \& When provide evaluation frameworks or taxonomies for agent failures~\citep{guo2026agenteval,cemri2025mast,zhang2025whoandwhen}; recent work further localizes agent failures to components, interactions, and points in a trajectory~\citep{raj2026modelharnessinteractioncentrictaxonomy,shah2026characterizingfaultsagenticai,zhao2026failureprocessanatomycli}. We build on this diagnostic line, but tie it directly to the artifact graph induced by a run. The graph lets us measure compute use, topology, coverage, monitoring, and selection in one object.

\section{Agentic Meta-Reasoning}
\label{sec:agentic-meta-reasoning}

Agentic meta-reasoning treats the choice of what to compute next as its own agentic reasoning task, separate from the object-level computation it directs. This disentangles the two cognitive functions of an agent while letting work accumulate without conditioning every control decision on its history.

\subsection{Memory, State, and Worker Context}
\label{sec:artifact-graphs}

An \emph{artifact} is a stored output from a worker or the controller. An attempt, a critique, or a controller's self-authored note can each be an artifact. Every artifact is stored with a stable identifier so it can be retrieved later. For a task \(x\), let \(M_t\) be the collection of artifacts available at control cycle \(t\).

The controller, at each cycle \(t\), updates its current self-authored textual state \(s_t\): a compact, mostly unstructured account of what the run has produced so far and what remains to be done. The state is rewritten each cycle, while the artifacts it references stay in memory. In the proof example, the memory may hold several artifacts that build up to the full argument. The controller's current state might record that one lemma is still unverified, and the controller can then launch parallel workers scoped to that sub-task. A worker receives the task, a controller-authored instruction \(g\), and as its context a set of artifacts \(C\subseteq M_t\) that the controller deems necessary for the assignment. If \(W\) denotes the worker execution and \(y\) its returned artifact, we write
\[
    y = W(x,g,C).
\]
For a proof check, the instruction says what to verify and the context supplies the existing arguments. Workers do not receive the controller's private state or deliberation, only the instructions and artifacts chosen for their assignment.

A worker can be a single model call or a coding agent that uses tools. The notation above describes an interface, not a pure function. A coding worker may read files not represented in \(C\), modify the environment, and produce stochastic outputs. The listed inputs therefore need not determine its output or its effects. We leave environment state and randomness implicit here, while recording the artifact context selected by the controller.

\subsection{Actions over the Run}
\label{sec:computation-actions}

The controller can inspect stored work, record its own notes, launch workers, or stop with an answer. These actions let it change both what work is done and what information is available for later decisions.

\paragraph{Read and write.}
For a collection of artifact identifiers \(I\), a memory function \(\mathrm{Read}(I)\) retrieves the corresponding artifacts. Similarly, for new memory content \(u\), \(\mathrm{Write}(u)\) stores it as an artifact with a fresh identifier. A note might preserve a warning that an earlier proof used an invalid assumption. These reads and writes are epistemic actions~\citep{kirsh1994distinguishing}: the controller leaves notes for itself to organize what it knows before acting on the task.

\paragraph{Run workers.}
To advance the object-level work, the controller can launch a batch of \(k\) parallel workers. For each worker \(i\), it supplies an instruction \(g_i\) and artifact context \(C_i\). Launching the batch applies the worker interface \(k\) times to the same task \(x\):
\[
    \mathrm{RunWorkers}\!\left(\{(g_i,C_i)\}_{i=1}^{k}\right)
    = \left\{\,W(x,g_i,C_i)\,\right\}_{i=1}^{k}.
\]
The workers' returned artifacts receive fresh identifiers and are added to memory, together with the identifiers of their input artifacts. The same interface supports a fresh attempt, a targeted repair, or a synthesis of earlier results. The assignment and context determine the kind of work and there are no fixed worker roles.

\paragraph{Stop and select.}
The controller ends the run with \(\mathrm{Stop}(y^{\star})\), where \(y^{\star}\) is an artifact already in memory that it selects as the final answer. It need not be the latest output, but it cannot be new: to submit a new artifact, the controller must first have a worker produce one.

\subsection{The Control Cycle}
\label{sec:staged-control}

Each control cycle has four sequential stages: \emph{Assess}, \emph{Propose}, \emph{Evaluate}, and \emph{Dispatch}. Each stage is an agentic process with its own prompts, context, and permitted memory operations and tools. A single stage can incorporate its own investigative actions and deliberation, so it may take several model calls to resolve. Between cycles, only a compact state is persisted and each stage may retrieve prior artifacts as needed. Memory and budget can change within a cycle as stages read, write, and consume model calls.

\paragraph{Assess: What have we learned?}
When workers return, the controller updates its running account of the run. In the proof example, it might record that the main argument looks promising but one lemma remains unverified. The components of the proof stay in memory; the state keeps track of what matters for the next decision. Let \(s_{t-1}\) be the previous assessment and \(\Delta M_t\) the newly available artifacts from the previous batch of workers. Assess produces the updated state \(s_t\):
\[
    s_t = \mathrm{Assess}(x,s_{t-1},\Delta M_t;M_t).
\]
The semicolon denotes access to persistent memory rather than inclusion of all its contents in the prompt.

\paragraph{Propose: What could we do next?}
Propose identifies next candidate computations from the current assessment to lay out the alternative lines of work explicitly before the committing to any of them. Propose is not given the remaining budget explicitly. This separates generating alternatives from judging their affordability, so potentially valuable but expensive options can enter the candidate set. Writing \(\mathcal{A}_t\) for the proposed options and recalling that \(s_t\) is the current assessment,
\[
    \mathcal{A}_t = \mathrm{Propose}(x,s_t;M_t).
\]

\paragraph{Evaluate: Which option is worth its cost?}
Evaluate weighs the proposed work from the previous stage against the remaining budget. If the proof hinges on one uncertain lemma, checking it may be more useful than starting another full attempt. If that check fails, the next cycle may favor a different approach. This is a prompted, qualitative assessment of computational value rather than an exact
optimization or a learned value-of-computation estimator.  Let \(b_t^{\mathrm{eval}}\) be the budget remaining when evaluation begins, after earlier calls have been charged. The selected proposal \(\tilde a_t\in\mathcal{A}_t\) is
\[
    \tilde a_t = \mathrm{Evaluate}(x,s_t,b_t^{\mathrm{eval}},\mathcal{A}_t;M_t).
\]

\paragraph{Dispatch: What should the worker receive?}
Dispatch turns the selected proposal into an executable action. For a lemma check, it writes the verification instruction and supplies the proof and any relevant critiques from artifact memory. A fresh attempt might receive no artifacts at all. Choosing which artifacts a worker sees is part of choosing the computation. Let \(a_t\) be the resulting worker-dispatch or stopping action. From the selected proposal \(\tilde a_t\), Dispatch produces
\[
    a_t = \mathrm{Dispatch}(x,s_t,\tilde a_t;M_t).
\]
If the decision is to stop, Dispatch issues \(\mathrm{Stop}(y^{\star})\) instead of launching workers. Otherwise, the returned worker outputs become available in the next cycle.

\subsection{Recording the Artifact Graph}
\label{sec:induced-graph}

For each worker, the dispatch stage selects prior artifacts as contexts which leaves a record of how later work builds on earlier work. If a worker receives a proof and returns a critique, the graph contains an edge from the proof to the critique. A repair that receives both artifacts has an incoming edge from each. Formally, let \(G=(V,E)\) be the artifact graph, with artifacts \(V\) and recorded dependency edges \(E\). For an artifact \(y_j\), let \(C_j\) be the earlier artifacts supplied as its context. Then
\[
    E = \{(y_i,y_j)\in V\times V: y_i\in C_j\}.
\]
Because the input artifacts already exist when the worker starts, these edges form a directed acyclic graph. Roots record work with no artifact inputs. Branches record multiple follow-ups, while several incoming edges show where a worker receives work from more than one source. In Section~\ref{sec:behavioral-analysis}, we examine these artifact graphs to characterize the computation that each run produces. We note that this induced graph, however, does not necessarily capture every information channel and only represents controller-generated dependencies. Where solution artifacts can be independently graded, we ask whether the graph ever contained a correct answer and whether that answer was submitted.

\subsection{Direct Control and the Cost of Deliberation}
\label{sec:direct-control}

We call the system that instantiates the design described so far the \textit{Meta-Reasoning Agent}. We also introduce a \textit{Direct Control Agent}, a matched variant with no explicit separation of control, but uses the same workers and the same interfaces for delegation, context selection, artifact writing, and stopping. It chooses each action in single turn over the accumulated history. The comparison thus isolates the combined control design, but it does not separately identify the effects of staging, state compression, or memory access.

Both systems receive the same nominal model-call budget \(B\). Let \(b_t\) be the allowance remaining at the start of cycle \(t\). If the cycle uses \(c_t\) controller calls and \(w_t\) worker calls, the next cycle starts with
\[
    b_{t+1} = b_t-c_t-w_t, \qquad b_0=B.
\]
Worker cost includes calls made inside a coding agent and sums calls across parallel workers. Controller cost includes every call within the four stages. If the budget is exhausted before the agent stops, the run is forced to submit an available answer. Note that equal call allowances do not imply equal token cost, latency, or FLOPs. Calls vary in length, and an agent may stop before using its allowance. Nor is spending more inherently better. The extra deliberation is useful only when it ultimately helps the agent produce a better solution.

\section{Diagnosing Agentic Inference Through Artifact Graphs}
\label{sec:behavioral-analysis}

Consider that in one run, none of the artifacts the workers produce is a correct proof. In a different run, a correct proof is already sitting in memory, but the controller submits a flawed revision instead. The performance score cannot tell these apart. The intermediate artifacts can separate the two cases, where the task admits a correctness label for the artifacts directly rather than only for the final submission. The measurements in this section are taken over those artifacts, asking how the run spent its budget and what structure that spending produced, whether a correct answer appeared at all, and whether the agent recognized it once it had.

\subsection{What Did the Agent Do with Its Budget?}

\paragraph{Did it use the available calls?}

An agent is free to stop before its budget allowance runs out, so a larger budget does not automatically convert to a longer run. Budget utilization measures the fraction of the allowance actually used. Let \(B\) be the nominal call budget, and let \(N_{\mathrm{ctrl}}\) and \(N_{\mathrm{work}}\) count controller and worker calls. Utilization is
\[
    U = \frac{N_{\mathrm{ctrl}}+N_{\mathrm{work}}}{B}.
\]
We read this measure alongside performance. Stopping early can be sensible, and spending every call can be wasteful. Together, usage and performance show whether a larger allowance leads to more work and better solutions.

\paragraph{Did later work build on earlier work?}
A run can spend its calls on independent attempts, or on building on, checking, and repairing earlier ones. The artifact graph distinguishes these patterns. Its nodes are worker outputs; its edges record which earlier outputs were supplied as context. We count nodes and edges to measure the amount of work and its recorded dependencies. Roots have no incoming edges; descendants have at least one. Fan-in counts a node's incoming edges, and we report its mean and maximum across nodes. Depth measures the longest dependency path in edges, with roots at depth zero. Width is the largest number of nodes at the same depth.

\paragraph{What information did the controller reuse?}
Memory reads and writes show which worker outputs and controller notes are stored and revisited. We examine what is read, how often it is read, and which stages perform these operations. We also measure the size of the agent's running state, the compact state written by the assess stage for meta-reasoning and the accumulating context for direct control.

\subsection{Did the Agent Find and Submit a Correct Answer?}

A run can succeed at finding the correct answer but still fail at recognizing and submitting it. We separate these two outcomes using correctness labels for intermediate solutions.

Let \(V_{\mathrm{sol}}\) be the artifacts that can be graded as candidate solutions. For a candidate \(y\), its label \(\ell(y)\) is one if it is correct and zero otherwise. The evaluated set must also include the submitted answer, denoted \(y^{\star}\). Let \(\mathcal{C}\) mean that the run contains a correct solution, and \(\mathcal{S}\) that its submission is correct. A correct submission must be among the run's solutions, so \(\mathcal{S}\subseteq\mathcal{C}\). Thus,
\[
    \Pr(\mathcal{S})
    = \Pr(\mathcal{C})\Pr(\mathcal{S}\mid\mathcal{C}).
\]
Success therefore factors into finding a correct answer and choosing it once one exists, and we call these terms \emph{coverage} and \emph{selection} respectively.

\paragraph{Coverage: Did a correct answer appear?}
Coverage counts a run as successful once it contains a correct solution, whether or not that solution is submitted. To track this over the run, let \(k\) be a call checkpoint. Let \(V_{\mathrm{sol},\leq k}\) contain the solution artifacts available within the first \(k\) calls, including controller and worker calls. Then
\[
    \mathrm{Coverage}(k)
    = \Pr\!\left(\exists y\in V_{\mathrm{sol},\leq k}:\ell(y)=1\right).
\]
This is the success rate a perfect selector could achieve from the answers already available. Low coverage means that many runs contain no correct candidate.

\paragraph{Monitoring: Could the agent recognize correct work?}
The worker that produced an artifact is the closest source of judgment about it, so we ask each worker to end its output with a confidence rating. We also ask the controller for its own judgment, which its assess stage records as a verdict on every candidate it reviews. Either signal is useful only if it ranks correct answers above incorrect ones. We evaluate the two signals separately, writing \(r(y)\) for whichever is under test on candidate \(y\). Draw a correct candidate \(Y^+\) and an incorrect candidate \(Y^-\) from the evaluated artifacts. We measure how often their scores put them in the right order, giving half credit for ties:
\[
    \mathrm{AUC}_2(r)
    = \Pr\!\left(r(Y^+)>r(Y^-)\right)
    +\frac{1}{2}\Pr\!\left(r(Y^+)=r(Y^-)\right).
\]
We call this Type-2 AUC because the system is judging its own answers~\citep{galvin2003type2,fleming2014measure} and it measures only discrimination, and not calibration. Note that the Direct Control Agent has no corresponding controller verdict signal, so we cannot make the same measurement for it.

\paragraph{Selection: Did it submit a correct answer it had found?}
We measure selection only among runs that contain a correct solution. Recalling that \(\mathcal{C}\) means a correct answer exists and \(\mathcal{S}\) means one was submitted,
\[
    \mathrm{Selection} = \Pr(\mathcal{S}\mid\mathcal{C}).
\]
A failed run is \emph{coverage-bound} if it never produced a correct solution. It is \emph{selection-bound} if it produced one but did not submit it.

\subsection{Does Selection Improve on a Structural Baseline?}

Consider a run arrives at the end with two proof candidates. One is correct and the other has a flaw. A uniform random choice between these candidates gives each an equal chance of being submitted; the better final decision should improve on that baseline.

We call the deepest terminal artifacts the \emph{convergence frontier}. For an artifact graph \(G\), let \(L(G)\) be its terminal nodes, those with no outgoing edges. Let \(d(y)\) be the longest path from a root to artifact \(y\), measured in edges. The frontier is
\[
    F(G) = \left\{y\in L(G):
    d(y)=\max_{z\in L(G)}d(z)\right\}.
\]
When the frontier artifacts have correctness labels, let \(q_F(G)\) be the probability that a uniform random choice from the frontier is correct. It is simply the fraction of frontier candidates with label \(\ell(y)=1\):
\[
    q_F(G)=\frac{1}{|F(G)|}\sum_{y\in F(G)}\ell(y).
\]
We compare this baseline with the correctness of the actual submission, \(\ell(y^{\star})\). Averaging their difference over runs that contain a correct solution, the event \(\mathcal{C}\), gives \emph{frontier selection gain}:
\[
    \mathrm{FSG}
    = \mathbb{E}\!\left[
        \ell(y^{\star})-q_F(G)
        \;\middle|\;\mathcal{C}
      \right].
\]
Positive gain means the agent does better than choosing uniformly from its frontier. The submission need not lie on that frontier, so the gain can also reflect choosing a better, shallower answer.

These diagnostics turn one final score into a more useful account of the run. They show what computation occurred, whether it produced correct work, and whether that work reached the final answer. The split also points at different remedies, since a coverage-bound failure motivates changes to exploration or object-level problem solving, while a selection-bound failure motivates tests of assessment, verification, and commitment.

\section{Experimental Setup}
\label{sec:experimental-setup}

We study whether agentic meta-reasoning improves final performance, whether its benefit grows with the available budget, and how it changes the work done during a run. We evaluate two different settings. On IMO ProofBench-Advanced, ARC-AGI-2, and LongCoT-mini, which we refer to together as the reasoning benchmarks, each worker is a model call that returns a text artifact and uses no external tools. On ProgramBench, each worker is a coding agent that can inspect files, run commands, and test a reconstruction.

\subsection{Tasks}
\label{subsec:tasks}

\paragraph{IMO ProofBench-Advanced.}

IMO ProofBench-Advanced~\citep{luong2025robust} contains 30 challenging proof-style problems. Candidate proofs are graded using the benchmark protocol from \citet{luong2025robust}; each solution receives a score in $\{0,1,6,7\}$. We sum the scores and report them as a percentage of the maximum possible total.

\paragraph{ARC-AGI-2.}
ARC-AGI-2~\citep{arcagi2} contains 120 abstract visual reasoning tasks posed as grids of colored cells. Each task shows a few input-output grid pairs that share one hidden transformation rule, and the solver must infer that rule and apply it to a held-out test input. Scoring is exact match on the output grid.

\paragraph{LongCoT-mini.}
LongCoT~\citep{longcot} contains long-horizon reasoning problems across logic, computer science, chemistry, chess, and mathematics; we evaluate its 507-problem mini split. Individual steps are tractable, and the difficulty lies in tracking state across many of them without dropping a constraint. Each problem has one verifiable answer, checked by a deterministic per-domain verifier.

\paragraph{ProgramBench.}
ProgramBench~\citep{yang2026programbench} contains 200 long-horizon program reconstruction tasks. Given documentation and an execute-only reference program, the solver must construct a self-contained codebase whose compiled executable matches the reference on hidden tests. The benchmark score is the fraction of hidden cases on which the candidate behavior matches the reference.

\subsection{Comparisons}

\paragraph{Meta-reasoning versus direct control.}
Our main comparison tests the Meta-Reasoning Agent against the Direct Control Agent. Both use the same workers. They share the same interfaces for delegation, artifact writing, context selection, and stopping. The Meta-Reasoning Agent uses compact state and staged control; the Direct Control Agent chooses each action in one call over full accumulated history. Neither controller is fine-tuned for these experiments.

\paragraph{Recursive Language Model.}
On the reasoning benchmarks, we also evaluate RLM~\citep{zhang2025recursive}. It places the task in a variable inside a Python interpreter rather than in the conversation. Each turn, the model writes code that calls a sub-model over part of that task, reads the printed result, and continues. A call can also spawn a child instance with its own turns, which is where the recursion arises. The agent has neither our worker-dispatch tool nor our artifact memory; its context is a single blob it edits with code. We make two modifications to the canonical design for comparability (Appendix~\ref{app:rlm}).

\paragraph{Coding agents.}
On ProgramBench we also compare against three complete coding agents: mini-SWE Agent~\citep{yang2024sweagent}, Claude Code~\citep{anthropic2026claudecode}, and Codex~\citep{openai2025codex}. We test mini-SWE Agent with all three models, and Codex and Claude Code with their native GPT-5.5 and Opus 4.8, in the same container environment our workers use (Appendix~\ref{app:coding}).
\subsection{Models and Budgets}

We use Gemini 3.1 Pro, GPT-5.5, and Opus 4.8 as the underlying models for all systems and benchmarks. The reasoning benchmarks use nominal budgets of 25, 50, and 100 model calls per problem. ProgramBench uses 400, 800, and 1200 calls, as a coding worker can make many calls during a single assignment. The main comparison uses the largest allowance in each setting: 100 calls for reasoning and 1200 for ProgramBench. External baselines receive the same nominal allowance for the comparison being reported.

We count every controller and worker model call toward the same budget. This includes all calls within controller stages and coding workers. Calls from parallel workers are summed, not counted as a single batch. Every system we evaluate is told how much of its allowance it has consumed and how much remains, and every one can stop early, so the calls a system actually uses may fall well short of its nominal allowance. The external baselines have no native notion of a model-call budget, so we add the same accounting to each (Appendices~\ref{app:rlm} and~\ref{app:coding}). If an agent exhausts it without stopping, the run is forced to submit an available solution.

\subsection{Measurements}

We report final performance using each benchmark's own metric, as defined in Section~\ref{subsec:tasks}. Budget sweeps show whether a larger allowance leads to better results.

To examine the work behind the scores, we use the diagnostics in Section~\ref{sec:behavioral-analysis}. We measure model calls used, artifact-graph structure, memory operations, and controller context. For the Meta-Reasoning Agent, we also report volume of memory operations by stage.

On the reasoning benchmarks we also grade intermediate solution artifacts as correct or incorrect, so that coverage, monitoring, selection, and frontier selection gain can be computed. We do not apply this analysis to ProgramBench, where intermediate reconstructions lack a corresponding grading protocol.

\section{Experimental Results}
\label{sec:results}

\subsection{Meta-Reasoning Improves Performance Across Tasks and Models}

\begin{figure}[t]
    \centering
    \includegraphics[width=\linewidth]{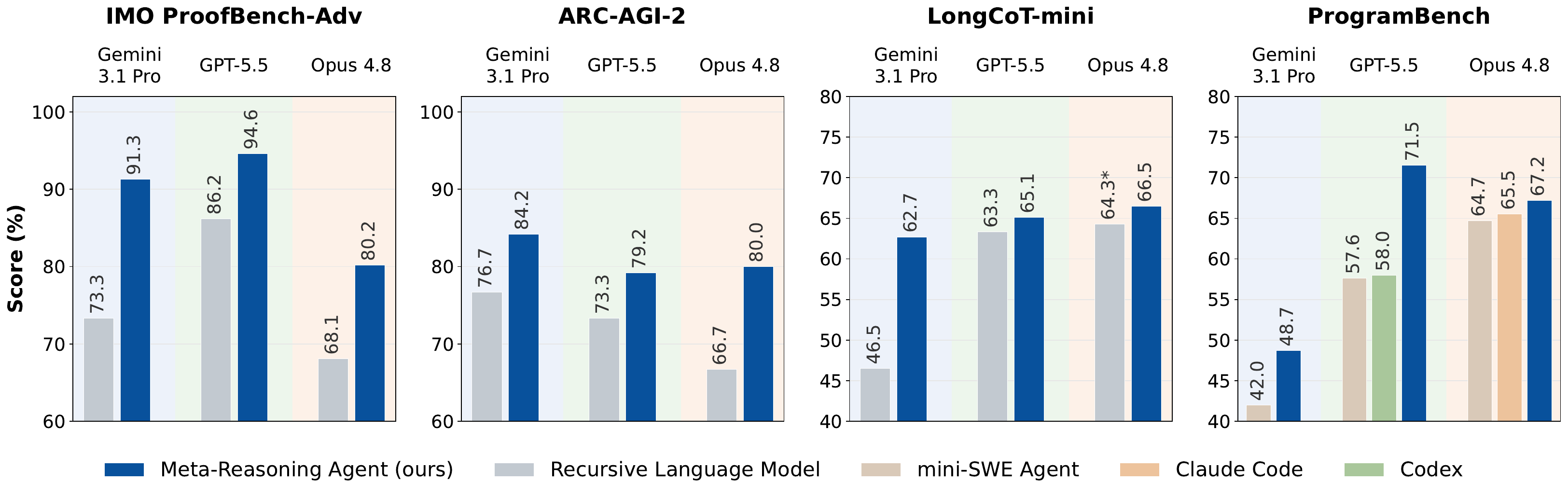}
    \caption{
    \textbf{Performance across four benchmarks and three models.}
    The Meta-Reasoning Agent is compared against external research
    harnesses and production coding agents. Nominal budgets are
    100 model calls on the reasoning benchmarks and 1200 on
    ProgramBench. Matched comparisons against Direct Control
    are reported in Table~\ref{tab:main}. An asterisk denotes
    RLM with the full Python workspace on LongCoT-mini
    (Appendix~\ref{app:rlm}).
    }
    \label{fig:main}
\end{figure}

\begin{table}[t]
\centering
\small
\begin{tabular}{llcccc}
\toprule
Model & System & \shortstack{IMO\\ProofBench-Adv} & ARC-AGI-2 & LongCoT-mini & ProgramBench \\
\midrule
Gemini 3.1 Pro & Recursive Language Model & 73.3 & 76.7 & 46.5 & -- \\
 & mini-SWE Agent & -- & -- & -- & 42.0 \\
 & Direct Control Agent & 82.7 & 77.5 & 53.5 & 46.9 \\
 & \textbf{Meta-Reasoning Agent} & \textbf{91.3} & \textbf{84.2} & \textbf{62.7} & \textbf{48.7} \\
\midrule
GPT-5.5 & Recursive Language Model & 86.2 & 73.3 & 63.3 & -- \\
 & mini-SWE Agent & -- & -- & -- & 57.6 \\
 & Codex & -- & -- & -- & 58.0 \\
 & Direct Control Agent & 93.3 & 75.8 & 64.7 & 63.7 \\
 & \textbf{Meta-Reasoning Agent} & \textbf{94.6} & \textbf{79.2} & \textbf{65.1} & \textbf{71.5} \\
\midrule
Opus 4.8 & Recursive Language Model & 68.1 & 66.7 & 64.3* & -- \\
 & mini-SWE Agent & -- & -- & -- & 64.7 \\
 & Claude Code & -- & -- & -- & 65.5 \\
 & Direct Control Agent & 78.0 & 77.5 & 65.3 & 65.3 \\
 & \textbf{Meta-Reasoning Agent} & \textbf{80.2} & \textbf{80.0} & \textbf{66.5} & \textbf{67.2} \\
\bottomrule
\end{tabular}
\caption{\textbf{Performance at the main nominal model-call budgets.} Budgets are 100 calls on the reasoning benchmarks and 1200 on ProgramBench, including controller calls. Values are percentages: normalized proof score on IMO ProofBench-Advanced, accuracy on ARC-AGI-2 and LongCoT-mini, and mean per-problem test-pass rate on ProgramBench. Bold marks the highest point estimate within each model and benchmark. A dash indicates an unreported comparison. An asterisk marks the full-Python RLM variant (Appendix~\ref{app:rlm}). External systems use the configurations described in Section~\ref{sec:experimental-setup}.}
\label{tab:main}
\end{table}

At the main budget settings, the Meta-Reasoning Agent scores highest in every model and benchmark pairing (Table~\ref{tab:main}, Figure~\ref{fig:main}). On IMO ProofBench-Advanced it improves by 4.0 points on average across the three models, with the largest gain, 8.6 points, coming with Gemini 3.1 Pro. ARC-AGI-2 improves by 4.2 points and is the most consistent, every model gaining between 2.5 and 6.7. LongCoT-mini improves by 3.6 points but has the highest spread, from 0.4 points with GPT-5.5 to 9.2 with Gemini 3.1 Pro, the largest single gain over direct control.

Meta-reasoning also achieves higher scores than the external
systems evaluated in each setting (Figure~\ref{fig:main}).
On ProgramBench with GPT-5.5, it reaches 71.5\%, compared
with 58.0\% for Codex. With Opus 4.8, it achieves 67.2\%,
compared with 65.5\% for Claude Code. It also exceeds
mini-SWE Agent with all three models, by 2.5--13.9 points.
These comparisons provide end-to-end reference points against
established systems, while the matched Direct Control comparison more directly tests the effect of changing how agentic computation is controlled.

\begin{findingbox}
At the main model-call budgets, meta-reasoning achieves higher
scores in all 12 matched comparisons and outperforms the
evaluated external agent configurations.
\end{findingbox}

\subsection{Meta-Reasoning Keeps Improving as the Budget Grows}

\begin{figure}[t]
  \centering
  \includegraphics[width=\linewidth]{
    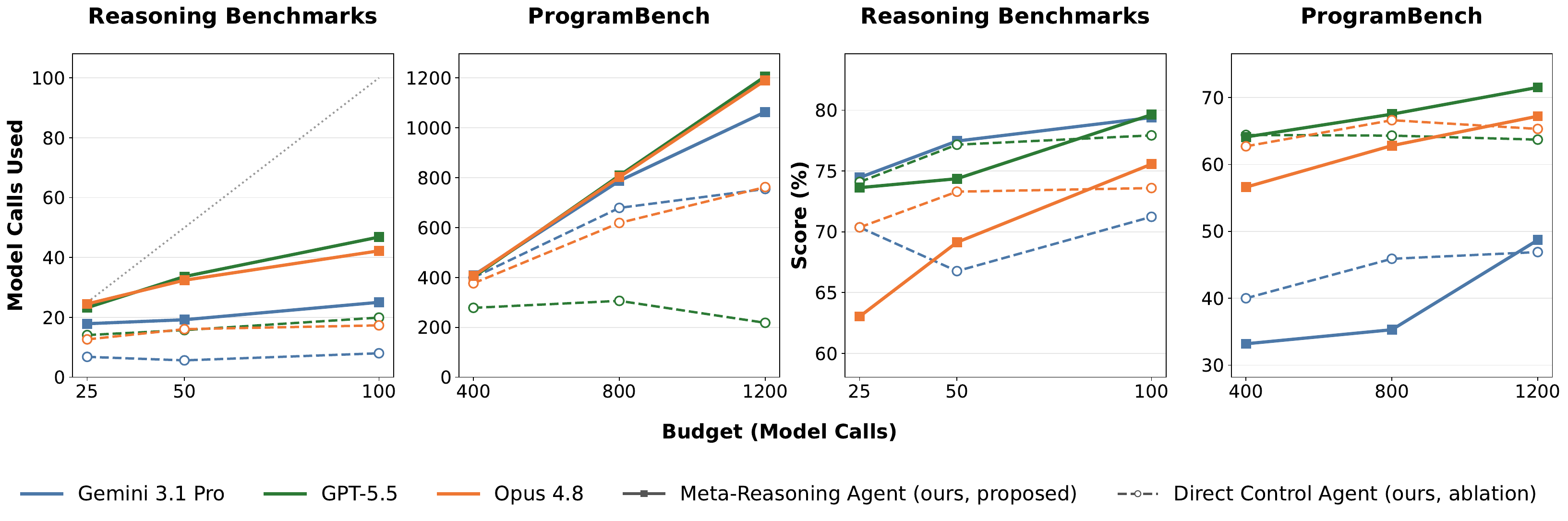}
  \caption{
    \textbf{Budget use and performance as the nominal allowance
    grows.}
    The left panels show actual model calls consumed, with a
    diagonal indicating full use of the allowance; the right
    panels show final scores. Reasoning results aggregate the
    three reasoning benchmarks; ProgramBench is separate
    because coding workers use many more calls.
  }
  \label{fig:scaling}
\end{figure}
A larger allowance helps only if the agent finds useful ways to spend it. Figure~\ref{fig:scaling} shows that meta-reasoning generally uses more calls as its budget grows. Direct control often stops early. On ProgramBench with GPT-5.5, it uses only about 18\% of the 1200-call allowance. Across models, meta-reasoning uses roughly 89\% to 101\% at this setting. The budget is enforced between control cycles, so workers dispatched while the run is still under the cap go to completion and can carry the total slightly past the allowance.

The longer runs also produce better answers. As the ProgramBench allowance grows from 400 to 1200 calls, meta-reasoning with GPT-5.5 rises from 64.1\% to 71.5\%. Direct control stays near 64\% across the sweep. The aggregate reasoning curves show the same contrast, with meta-reasoning continuing to improve over the tested range while direct control more often plateaus.

Stopping early is not the whole explanation. With Opus 4.8 on ProgramBench, direct control roughly doubles its calls used over the sweep, from 376 to 768. Its score rises only modestly, from 62.7\% to 65.3\%, and peaks at the intermediate budget. Continuing to work is not enough; what the agent does next also matters.

Meta-reasoning does not win at every budget. With Opus 4.8, it trails direct control at 400 calls, 56.6\% versus 62.7\%, then leads at 1200 calls, 67.2\% versus 65.3\%. GPT-5.5 is also slightly behind at the smallest ProgramBench allowance before pulling ahead. This is consistent with a cost to staged control that takes time to repay.

\begin{findingbox}
Meta-reasoning continues improving over the tested budget ranges where direct control often plateaus. At smaller allowances, overhead from explicit control stages can outweigh its benefit.
\end{findingbox}

\subsection{What computation does the additional budget produce?}

\begin{figure}[t]
    \centering
    \includegraphics[width=\linewidth]{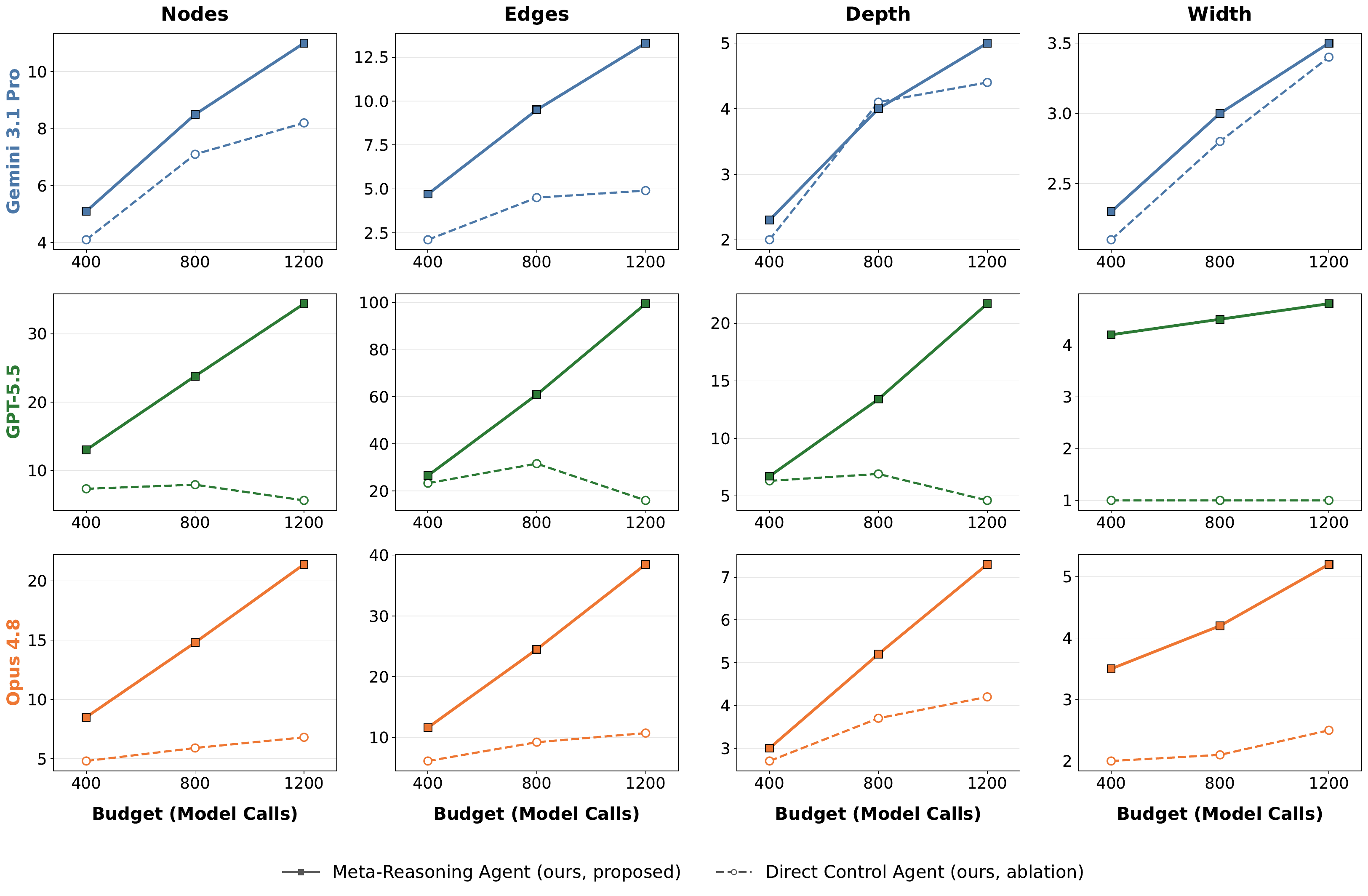}
    \caption{\textbf{Worker-artifact topology on ProgramBench across nominal budgets.} Nodes are worker outputs and edges are recorded dependencies through selected artifact context. Depth is the longest dependency path and width is the largest number of worker artifacts at a common depth.}
    \label{fig:topology}
\end{figure}

\begin{figure}[t]
    \centering
    \includegraphics[width=\linewidth]{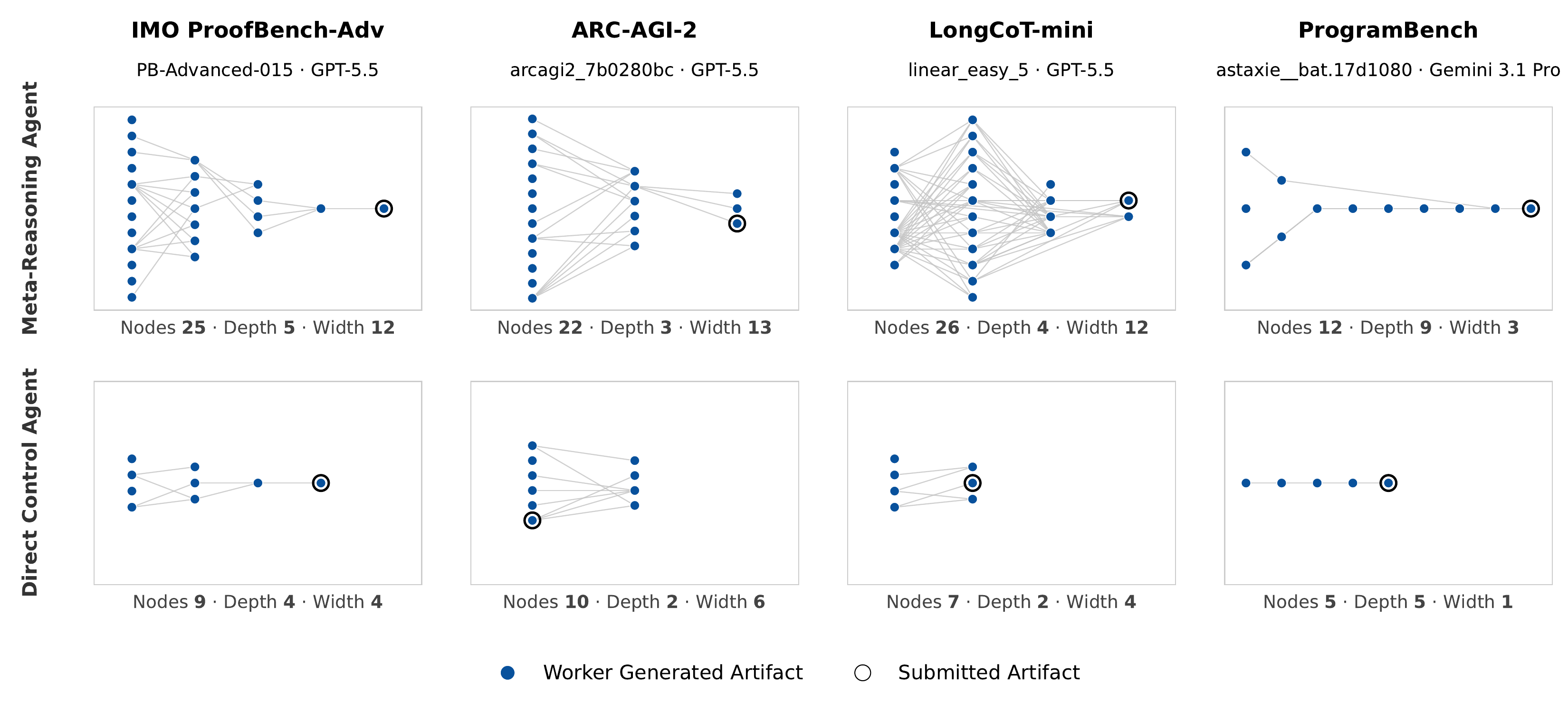}
    \caption{\textbf{Example artifact graphs from meta-reasoning and direct control runs on matched problems.} Edges indicate recorded selected-context dependencies; the ringed node denotes the submitted artifact.}
    \label{fig:dags}
\end{figure}

Meta-reasoning produces more worker outputs than direct control on all three reasoning benchmarks at the main budget, and the reuse between them grow faster still. For instance, on IMO ProofBench-Advanced with Gemini 3.1 Pro, worker artifacts roughly double while recorded dependencies rise by an order of magnitude. Both agents select which earlier artifacts each worker receives through the same interface, so direct control could build the same graph but largely does not.

Graph size also grows with the budget. On ProgramBench with GPT-5.5, raising the allowance from 400 to 1200 calls produces more than 2.5 times as many nodes and nearly four times as many edges (Figure~\ref{fig:topology}). Direct control's graphs show no comparable growth. At the largest allowance, meta-reasoning produces more nodes and edges than direct control with each model.

Figure~\ref{fig:dags} shows these structural differences on matched problems from each benchmark. For example, in the ARC-AGI-2 run, direct control generates six independent attempts and four one-hop follow-ups, while meta-reasoning explores more independent attempts, connects later work to earlier results, and reaches a frontier of three candidates. It combines exploration and reuse within the same run.

\begin{findingbox}
As the budget grows, meta-reasoning produces more worker outputs and builds more of them on earlier ones.
\end{findingbox}

\subsection{Finding Correct Answers Is Only Part of the Problem}

\begin{figure}[t]
    \centering
    \includegraphics[width=\linewidth]{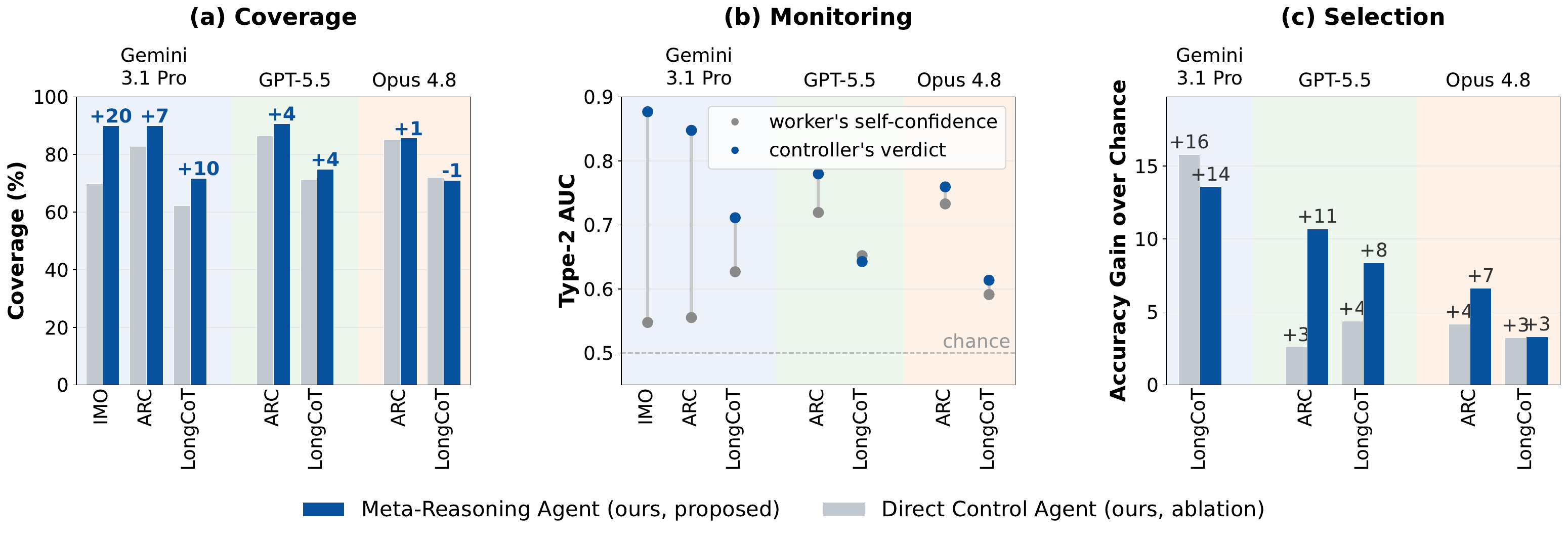}
    \caption{
    \textbf{Finding, monitoring, and selecting correct work.}
    (a) Correct-candidate coverage;
    (b) Type-2 AUC for worker confidence and, for
    meta-reasoning only, Assess-stage verdicts;
    (c) frontier selection gain relative to a uniform choice
    from each run's own deepest terminal artifacts.
    The measures and their scopes are defined in
    Section~\ref{sec:behavioral-analysis}. Coverage on IMO
    ProofBench-Advanced uses binary correctness, not the
    partial-credit proof score.
    }
    \label{fig:behavior}
\end{figure}

\paragraph{Correct answers appear in more runs.}
Meta-reasoning improves coverage in most evaluated settings (Figure~\ref{fig:behavior}(a)). The largest gains occur with Gemini 3.1 Pro: 20 percentage points on IMO ProofBench-Advanced and 10 on LongCoT-mini. ARC-AGI-2 improves across models, while Opus 4.8 on LongCoT-mini is approximately tied. These gains mean that more runs contain a correct candidate, whether or not it is submitted as the final answer.

\paragraph{The controller often judges quality better than worker confidence does.}
Worker confidence can be a weak guide to correctness. For Gemini 3.1 Pro on IMO ProofBench-Advanced, its ranking is near chance, with a Type-2 AUC of 0.55. The controller's verdict reaches 0.88 (Figure~\ref{fig:behavior}(b)). Gemini 3.1 Pro on ARC-AGI-2 shows a similarly large gap. The difference is smaller with GPT-5.5 and Opus 4.8, and nearly absent for GPT-5.5 on LongCoT-mini. Explicit controller-level assessment therefore gives a stronger ranking signal than worker confidence in several settings. Whether that signal reaches the final answer is the next question.

\paragraph{The final choice improves in some settings, but not all.}
Agents usually submit an answer from the deepest terminal artifacts, which we call the convergence frontier. This happens in 83\% of the ARC-AGI-2 and LongCoT-mini runs summarized in Figure~\ref{fig:behavior}(c). Yet roughly three-quarters of these runs have several candidates at that frontier, so the agent still has to pick which one to submit.

Frontier selection gain compares that submission with a random choice from the same frontier. As Figure~\ref{fig:behavior}(c) shows, on ARC-AGI-2 with GPT-5.5 meta-reasoning gains 11 percentage points, compared with 3 for direct control. Meta-reasoning also has higher gain on ARC-AGI-2 with Opus 4.8 and LongCoT-mini with GPT-5.5. On LongCoT-mini, the agents are approximately tied for Opus 4.8, while direct control is slightly ahead for Gemini 3.1 Pro.

\begin{findingbox}
For an agent, finding the correct answer and submitting it are different challenges. Meta-reasoning improves coverage in most evaluated settings, but its gains in frontier selection are not uniform across models.
\end{findingbox}

\subsection{Can control retain useful history without
replaying it all?}

\begin{figure}[t]
    \centering
    \includegraphics[width=\linewidth]{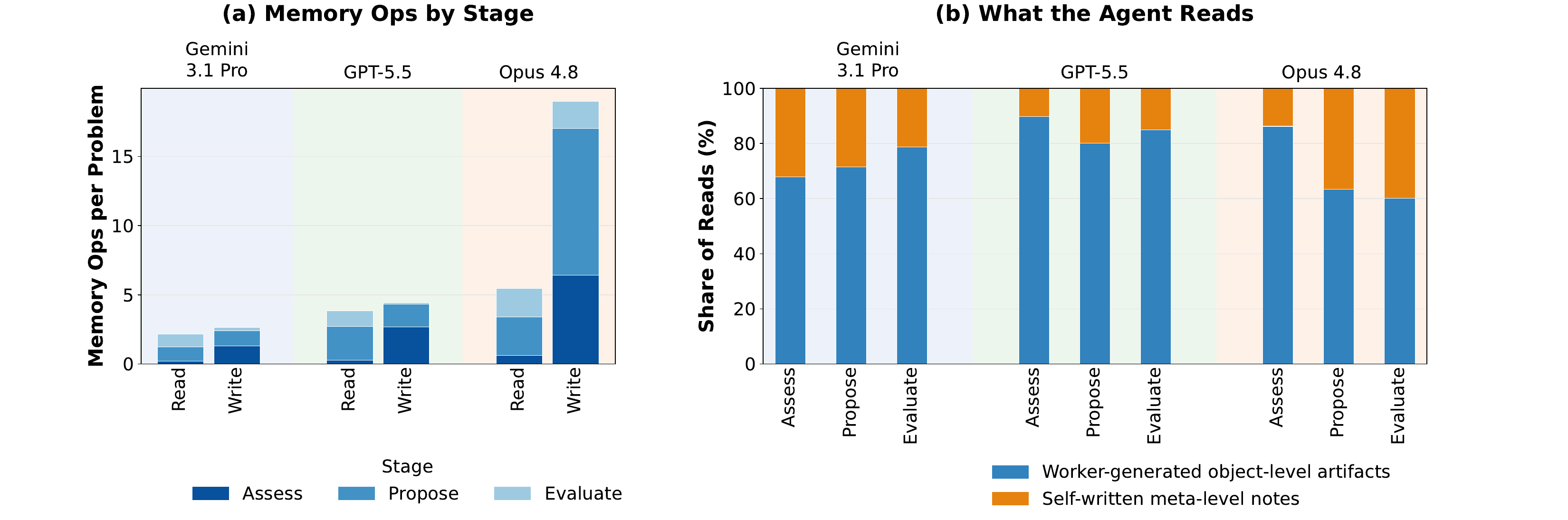}
    \caption{\textbf{Meta-Reasoning Agent's memory use on ARC-AGI-2 and LongCoT-mini.} Operations are grouped by stage, and reads are separated by whether they target worker outputs or controller-authored notes.}
    \label{fig:memory}
\end{figure}

\begin{figure}[t]
    \centering
    \includegraphics[width=\linewidth]{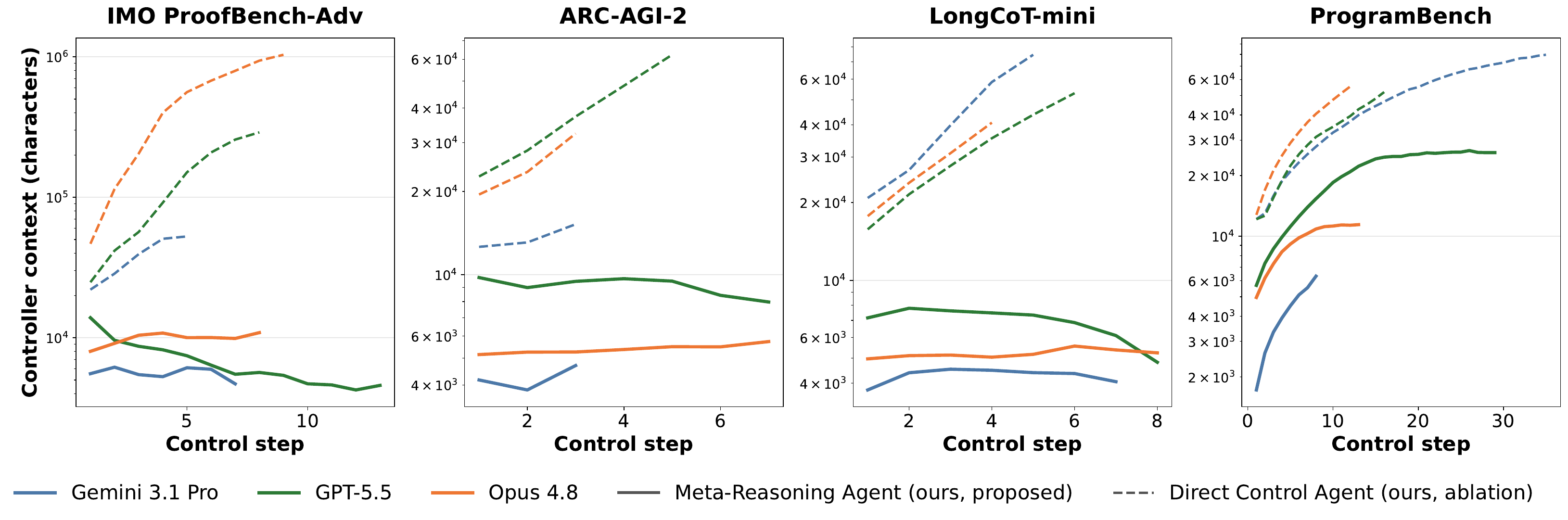}
    \caption{
    \textbf{Size of the representation persisted between
    control decisions (log scale).}
    For the Direct Control Agent the representation is the accumulated message history; for the Meta-Reasoning Agent it is the compact state maintained by its assess stage.
    }
    \label{fig:context}
\end{figure}

Artifact memory holds worker outputs and the controller's own notes side by side. We observe that the meta-reasoning controller re-reads its own notes more intensively than worker outputs (Figure~\ref{fig:memory}). Among artifacts explicitly retrieved at least once, notes average 4.4 to 9.8 reads on ARC-AGI-2 and LongCoT-mini, against 1.9 to 3.3 for worker artifacts. Many artifacts are never retrieved again, but interestingly the controller revisits a small portion of its own notes repeatedly, using them as recurring organizational references.

Memory operations concentrate in assess and propose (Figure~\ref{fig:memory}(a)). Writes outnumber reads for every model. Assess writes steadily but issues almost no reads, which follows from its inputs: new worker artifacts arrive in its prompt, so an explicit read is a request for older work instead. Propose and evaluate account for nearly all the reads, drawing on what the run has already produced. Memory use also rises sharply across models: Gemini 3.1 Pro uses it lightly, GPT-5.5 more often, and Opus 4.8 writes many notes that are never revisited.

A major advantage of persistent artifact storage is that it enables the controller to maintain a smaller account of the run instead of the full interaction history. Direct control's history exceeds one million characters on the longest Opus 4.8 IMO ProofBench-Advanced runs (Figure~\ref{fig:context}). The meta-reasoning state stays on the order of thousands to tens of thousands of characters. On the reasoning benchmarks, it often shrinks as the run progresses. On ProgramBench the gap narrows to a factor of a few, and the meta-reasoning state grows over the run rather than staying flat, since much of what the controller reads there is repository content it must keep an account of.

Direct control's history reaches a size at which models are known to have trouble locating relevant information~\citep{liu2023lostmiddlelanguagemodels}. This may be part of why direct control stops improving as its budget grows, though we did not test it directly and isolating the effect would require a separate ablation. We note that these counts measure the controller's state alone, so they set a floor on what each decision actually sees. Retrieved artifacts and within-stage interactions add further context.

\begin{findingbox}
The meta-reasoning controller decouples the size of its running state from the length of the run while selectively revisiting its own notes as recurring organizational references.
\end{findingbox}

\section{Discussion and Limitations}
\label{sec:discussion}

\subsection{Meta-Reasoning Is Part of Agent Capability}

How well an agent performs on a long task is usually read as a property of the model doing the work. Our comparison holds that model fixed. The Meta-Reasoning Agent and the Direct Control Agent call the same models, dispatch the same workers, and draw on the same allowance, differing only in how the run is directed. We observe that every matched setting improves, and the margin widens as the budget allowance grows. The ability to direct a run is therefore a component of agent performance in its own right, one that can be measured and improved separately from the work being directed.

\subsection{Meta-Reasoning Is Itself an Agentic Process}

An agent's meta-level reasoning is usually interleaved with the object-level work, sharing the same call, the same context, and the same accumulating history. What we argue conceptually, and observe empirically, is that this level of cognition can itself be a different kind of work, one that takes multiple steps, dedicated tools, and sub-tasks of its own. Before committing to a next computation, the controller retrieves earlier artifacts, writes notes for itself and returns to a few of them repeatedly, and records judgments on candidates that rank correctness better than the workers' own confidence does. These are epistemic actions in the sense of \citet{kirsh1994distinguishing}, moves whose purpose is to change what the agent knows rather than to advance the task directly. The persistent memory participates in that reasoning, holding what the controller deems note-worthy so that later stages can act on it, in the way external scaffolding has been argued to participate in cognition rather than merely record it~\citep{clarkemt}.

The four stages we propose are one way to organize that activity. What the results support is the separation of these different functions. Judging a growing body of interconnected work is itself a complex problem, so the meta level needs the same affordances as the object level, somewhere to keep what it has worked out, the ability to go and look before deciding, and more than one step in which to do so.

\subsection{Limitations}

\paragraph{Costs and failure modes.}
Our proposed staged control costs additional compute. The low-budget crossovers in Figure~\ref{fig:scaling} show that the design is not uniformly preferable to direct control. Additionally, an incorrect controller assessment can also propagate a misleading state or discard good partial work it never re-reads, and compact state is lossy, relying on future memory reads to surface information that becomes important only later.

An interesting failure case we observe comes from the LongCoT-mini chess subset. Here, meta-reasoning underperforms direct control while outperforming in all other domains by a large margin. Our preliminary sampling of trajectories points to unproductive reconsideration, in which additional checking or elaboration destabilizes an answer that was already correct. We leave full analysis and targeted interventions for future work.

\paragraph{Model dependence.}

In our experiments, the size of the gain also depends on the model used. Gemini 3.1 Pro benefits strongly on the reasoning benchmarks, GPT-5.5 especially on ProgramBench, and Opus 4.8 shows smaller but uniformly positive gains across tasks. These differences describe how each model behaves as a controller in our harness, where task difficulty, prompting, and worker competence all enter.

Appendix~\ref{app:scope} sets out the full scope of the evidence.

\section{Conclusion}
\label{sec:conclusion}

An agent given more inference-time computation faces more choices about what to do with it. We introduced agentic meta-reasoning to make those choices an explicit part of the agent. Across four benchmarks and three frontier models, the Meta-Reasoning Agent improves every matched comparison at the main budget settings and keeps improving over the evaluated budget ranges where direct control often plateaus. Our analysis shows that it spends more of its allowance, composes later work out of earlier results, produces a correct candidate in more runs, and judges its own work more reliably than worker confidence does, while decoupling the size of its state from the length of the run.

An agent's performance therefore depends on how well it decides what work is worth doing, alongside how well it does the work. As runs grow longer, that decision becomes work in its own right.

\bibliographystyle{plainnat}
\bibliography{paper_latest}

\begin{thebibliography}{58}
\providecommand{\natexlab}[1]{#1}
\providecommand{\url}[1]{\texttt{#1}}
\expandafter\ifx\csname urlstyle\endcsname\relax
  \providecommand{\doi}[1]{doi: #1}\else
  \providecommand{\doi}{doi: \begingroup \urlstyle{rm}\Url}\fi

\bibitem[{Anthropic}(2026)]{anthropic2026claudecode}
{Anthropic}.
\newblock Claude code, 2026.
\newblock \url{https://www.anthropic.com/product/claude-code}.

\bibitem[Bertsch et~al.(2025)Bertsch, Pratapa, Mitamura, Neubig, and
  Gormley]{bertsch2025oolongevaluatinglongcontext}
Amanda Bertsch, Adithya Pratapa, Teruko Mitamura, Graham Neubig, and Matthew~R.
  Gormley.
\newblock Oolong: Evaluating long context reasoning and aggregation
  capabilities, 2025.
\newblock \url{https://arxiv.org/abs/2511.02817}.

\bibitem[Besta et~al.(2024)Besta, Blach, Kubicek, Gerstenberger, Podstawski,
  Gianinazzi, Gajda, Lehmann, Niewiadomski, Nyczyk, and
  Hoefler]{besta2024graph}
Maciej Besta, Nils Blach, Ales Kubicek, Robert Gerstenberger, Michal
  Podstawski, Lukas Gianinazzi, Joanna Gajda, Tomasz Lehmann, Hubert
  Niewiadomski, Piotr Nyczyk, and Torsten Hoefler.
\newblock Graph of thoughts: Solving elaborate problems with large language
  models.
\newblock \emph{Proceedings of the AAAI Conference on Artificial Intelligence},
  38\penalty0 (16):\penalty0 17682--17690, 2024.

\bibitem[Brown et~al.(2024)Brown, Juravsky, Ehrlich, et~al.]{brown2024monkeys}
Bradley Brown, Jordan Juravsky, Ryan Ehrlich, et~al.
\newblock Large language monkeys: Scaling inference compute with repeated
  sampling, 2024.
\newblock \url{https://arxiv.org/abs/2407.21787}.

\bibitem[Cao et~al.(2026)Cao, Wang, Qin, et~al.]{cao2026knowact}
Qi~Cao, Yufan Wang, Peijia Qin, et~al.
\newblock Llms know when they know, but do not act on it: A metacognitive
  harness for test-time scaling, 2026.
\newblock \url{https://arxiv.org/abs/2605.14186}.

\bibitem[Cemri et~al.(2025)Cemri, Pan, Yang, et~al.]{cemri2025mast}
Mert Cemri, Melissa~Z. Pan, Shuyi Yang, et~al.
\newblock Why do multi-agent llm systems fail?
\newblock In \emph{Advances in Neural Information Processing Systems (NeurIPS),
  Datasets and Benchmarks Track}, 2025.
\newblock \url{https://arxiv.org/abs/2503.13657}.

\bibitem[Chhikara et~al.(2025)Chhikara, Khant, Aryan, et~al.]{chhikara2025mem0}
Prateek Chhikara, Dev Khant, Saket Aryan, et~al.
\newblock Mem0: Building production-ready ai agents with scalable long-term
  memory, 2025.
\newblock \url{https://arxiv.org/abs/2504.19413}.

\bibitem[Chollet et~al.(2026)Chollet, Knoop, Kamradt, Landers, and
  Pinkard]{arcagi2}
Francois Chollet, Mike Knoop, Gregory Kamradt, Bryan Landers, and Henry
  Pinkard.
\newblock Arc-agi-2: A new challenge for frontier ai reasoning systems, 2026.
\newblock \url{https://arxiv.org/abs/2505.11831}.

\bibitem[Clark and Chalmers(1998)]{clarkemt}
Andy Clark and David Chalmers.
\newblock The extended mind.
\newblock \emph{Analysis}, 58\penalty0 (1):\penalty0 7--19, 1998.
\newblock ISSN 00032638, 14678284.
\newblock \url{http://www.jstor.org/stable/3328150}.

\bibitem[De~Sabbata et~al.(2024)De~Sabbata, Sumers, AlKhamissi,
  et~al.]{desabbata2024rationalmetareasoning}
C.~Nicol{\`o} De~Sabbata, Theodore~R. Sumers, Badr AlKhamissi, et~al.
\newblock Rational metareasoning for large language models, 2024.
\newblock \url{https://arxiv.org/abs/2410.05563}.

\bibitem[Dong et~al.(2025)Dong, Ye, Zhu, et~al.]{dong2025metar1}
Haonan Dong, Haoran Ye, Wenhao Zhu, et~al.
\newblock Meta-r1: Empowering large reasoning models with metacognition, 2025.
\newblock \url{https://arxiv.org/abs/2508.17291}.

\bibitem[Dragoi et~al.(2026)Dragoi, Pintilie, Gogianu, and
  Brad]{dragoi2025beyondpassk}
Marius Dragoi, Ioana Pintilie, Florin Gogianu, and Florin Brad.
\newblock Beyond pass@k: Breadth-depth metrics for reasoning boundaries.
\newblock In \emph{International Conference on Learning Representations
  (ICLR)}, 2026.
\newblock \url{https://arxiv.org/abs/2510.08325}.

\bibitem[Fleming and Lau(2014)]{fleming2014measure}
Stephen~M. Fleming and Hakwan~C. Lau.
\newblock How to measure metacognition.
\newblock \emph{Frontiers in Human Neuroscience}, 8:\penalty0 443, 2014.
\newblock \doi{10.3389/fnhum.2014.00443}.
\newblock \url{https://doi.org/10.3389/fnhum.2014.00443}.

\bibitem[Fox et~al.(2026)Fox, Wang, Rosu, and
  Dhingra]{fox2026prolongprogrammaticmemoryenables}
Alexis Fox, Junlin Wang, Paul Rosu, and Bhuwan Dhingra.
\newblock Pro-long: Programmatic memory enables long-horizon reasoning, 2026.
\newblock \url{https://arxiv.org/abs/2607.20064}.

\bibitem[Galvin et~al.(2003)Galvin, Podd, Drga, and Whitmore]{galvin2003type2}
Susan~J. Galvin, John~V. Podd, Vit Drga, and John Whitmore.
\newblock Type 2 tasks in the theory of signal detectability: Discrimination
  between correct and incorrect decisions.
\newblock \emph{Psychonomic Bulletin \& Review}, 10\penalty0 (4):\penalty0
  843--876, 2003.
\newblock \doi{10.3758/BF03196546}.

\bibitem[Guo et~al.(2026)Guo, Wu, and Yiu]{guo2026agenteval}
Wei Guo, Yuchen Wu, and Sui-Ming Yiu.
\newblock Agenteval: Dag-structured step-level evaluation for agentic workflows
  with error propagation tracking.
\newblock In \emph{Proceedings of ACL 2026 (Industry Track)}, 2026.
\newblock \url{https://arxiv.org/abs/2604.23581}.

\bibitem[Kim et~al.(2024)Kim, Moon, Tabrizi, et~al.]{kim2024llmcompiler}
Sehoon Kim, Suhong Moon, Ryan Tabrizi, et~al.
\newblock An llm compiler for parallel function calling.
\newblock In \emph{International Conference on Machine Learning (ICML)}, 2024.
\newblock \url{https://arxiv.org/abs/2312.04511}.

\bibitem[Kirsh and Maglio(1994)]{kirsh1994distinguishing}
David Kirsh and Paul Maglio.
\newblock On distinguishing epistemic from pragmatic action.
\newblock \emph{Cognitive Science}, 18\penalty0 (4):\penalty0 513--549, 1994.
\newblock ISSN 0364-0213.
\newblock \doi{10.1016/0364-0213(94)90007-8}.
\newblock
  \url{https://www.sciencedirect.com/science/article/pii/0364021394900078}.

\bibitem[Lee et~al.(2026)Lee, Nair, Zhang, Lee, Khattab, and
  Finn]{lee2026metaharness}
Yoonho Lee, Roshen Nair, Qizheng Zhang, Kangwook Lee, Omar Khattab, and Chelsea
  Finn.
\newblock Meta-harness: End-to-end optimization of model harnesses, 2026.
\newblock \url{https://arxiv.org/abs/2603.28052}.

\bibitem[Li et~al.(2025)Li, Li, Dong, et~al.]{li2025meco}
Wenjun Li, Dexun Li, Kuicai Dong, et~al.
\newblock Adaptive tool use in large language models with meta-cognition
  trigger.
\newblock In \emph{Proceedings of ACL 2025}, 2025.
\newblock \url{https://arxiv.org/abs/2502.12961}.

\bibitem[Lin et~al.(2026)Lin, Liu, Pan, Lin, Dou, Xi, Huang, Yan, Han, Gui, and
  Jiang]{lin2026agenticharnessengineeringobservabilitydriven}
Jiahang Lin, Shichun Liu, Chengjun Pan, Lizhi Lin, Shihan Dou, Zhiheng Xi,
  Xuanjing Huang, Hang Yan, Zhenhua Han, Tao Gui, and Yu-Gang Jiang.
\newblock Agentic harness engineering: Observability-driven automatic evolution
  of coding-agent harnesses, 2026.
\newblock \url{https://arxiv.org/abs/2604.25850}.

\bibitem[Liu et~al.(2026)Liu, Gani, Lu, Thomas, Steyvers, and
  Cohan]{liu2026metacognitionsurvey}
Gabrielle Kaili-May Liu, Areeb Gani, Jacqueline Lu, Jordan Thomas, Mark
  Steyvers, and Arman Cohan.
\newblock Metacognition in llms: Foundations, progress, and opportunities,
  2026.
\newblock \url{https://arxiv.org/abs/2607.11881}.

\bibitem[Liu et~al.(2023)Liu, Lin, Hewitt, Paranjape, Bevilacqua, Petroni, and
  Liang]{liu2023lostmiddlelanguagemodels}
Nelson~F. Liu, Kevin Lin, John Hewitt, Ashwin Paranjape, Michele Bevilacqua,
  Fabio Petroni, and Percy Liang.
\newblock Lost in the middle: How language models use long contexts, 2023.
\newblock \url{https://arxiv.org/abs/2307.03172}.

\bibitem[Lou et~al.(2026)Lou, Lázaro-Gredilla, Dedieu, Wendelken, Lehrach, and
  Murphy]{lou2026autoharnessimprovingllmagents}
Xinghua Lou, Miguel Lázaro-Gredilla, Antoine Dedieu, Carter Wendelken,
  Wolfgang Lehrach, and Kevin~P. Murphy.
\newblock Autoharness: improving llm agents by automatically synthesizing a
  code harness, 2026.
\newblock \url{https://arxiv.org/abs/2603.03329}.

\bibitem[Luong et~al.(2025)Luong, Hwang, Nguyen, Ghiasi, Chervonyi, Seo, Kim,
  Bingham, Lee, Mishra, Zhai, Hu, Michalewski, Kim, Ahn, Bae, Song, Trinh, Le,
  and Jung]{luong2025robust}
Thang Luong, Dawsen Hwang, Hoang~H. Nguyen, Golnaz Ghiasi, Yuri Chervonyi,
  Insuk Seo, Junsu Kim, Garrett Bingham, Jonathan Lee, Swaroop Mishra, Alex
  Zhai, Clara~Huiyi Hu, Henryk Michalewski, Jimin Kim, Jeonghyun Ahn, Junhwi
  Bae, Xingyou Song, Trieu~H. Trinh, Quoc~V. Le, and Junehyuk Jung.
\newblock Towards robust mathematical reasoning.
\newblock In \emph{Proceedings of the 2025 Conference on Empirical Methods in
  Natural Language Processing}, 2025.
\newblock \url{https://aclanthology.org/2025.emnlp-main.1794/}.

\bibitem[Madaan et~al.(2023)Madaan, Tandon, Gupta,
  et~al.]{madaan2023selfrefine}
Aman Madaan, Niket Tandon, Prakhar Gupta, et~al.
\newblock Self-refine: Iterative refinement with self-feedback.
\newblock In \emph{Advances in Neural Information Processing Systems
  (NeurIPS)}, 2023.
\newblock \url{https://arxiv.org/abs/2303.17651}.

\bibitem[Motwani et~al.(2026)Motwani, Nichols, London, Li, Pizzati, Blake,
  Hammoud, McDonald, Naik, Ivanova, Baskaran, Laptev, Glatt, Ben-Nun, Torr,
  Jaques, Prabhu, Bartoldson, Kailkhura, and de~Witt]{longcot}
Sumeet~Ramesh Motwani, Daniel Nichols, Charles London, Peggy Li, Fabio Pizzati,
  Acer Blake, Hasan Hammoud, Tavish McDonald, Akshat Naik, Alesia Ivanova,
  Vignesh Baskaran, Ivan Laptev, Ruben Glatt, Tal Ben-Nun, Philip Torr, Natasha
  Jaques, Ameya Prabhu, Brian Bartoldson, Bhavya Kailkhura, and
  Christian~Schroeder de~Witt.
\newblock Longcot: Benchmarking long-horizon chain-of-thought reasoning, 2026.
\newblock \url{https://arxiv.org/abs/2604.14140}.

\bibitem[Nielsen et~al.(2026)Nielsen, Cetin, Schwendeman,
  et~al.]{nielsen2025conductor}
Stefan Nielsen, Edoardo Cetin, Peter Schwendeman, et~al.
\newblock Learning to orchestrate agents in natural language with the
  conductor.
\newblock In \emph{International Conference on Learning Representations
  (ICLR)}, 2026.
\newblock \url{https://arxiv.org/abs/2512.04388}.

\bibitem[Oh and Gobet(2025)]{oh2025mgv}
Nick Oh and Fernand Gobet.
\newblock Monitor-generate-verify: Formalising metacognitive theory for
  language model reasoning, 2025.
\newblock \url{https://arxiv.org/abs/2511.04341}.

\bibitem[{OpenAI}(2025)]{openai2025codex}
{OpenAI}.
\newblock Introducing codex, May 2025.
\newblock \url{https://openai.com/index/introducing-codex/}.

\bibitem[Packer et~al.(2023)Packer, Wooders, Lin, et~al.]{packer2023memgpt}
Charles Packer, Sarah Wooders, Kevin Lin, et~al.
\newblock Memgpt: Towards llms as operating systems, 2023.
\newblock \url{https://arxiv.org/abs/2310.08560}.

\bibitem[Prasad et~al.(2024)Prasad, Koller, Hartmann, et~al.]{prasad2024adapt}
Archiki Prasad, Alexander Koller, Mareike Hartmann, et~al.
\newblock Adapt: As-needed decomposition and planning with language models.
\newblock In \emph{Findings of NAACL 2024}, 2024.
\newblock \url{https://arxiv.org/abs/2311.05772}.

\bibitem[Raj et~al.(2026)Raj, Gupta, Mahmoud, Dumitru, Yi, Sabharwal, and
  He]{raj2026modelharnessinteractioncentrictaxonomy}
Harsh Raj, Vipul Gupta, Anas Mahmoud, Razvan-Gabriel Dumitru, Darvin Yi, Aakash
  Sabharwal, and Yunzhong He.
\newblock Model or harness? an interaction-centric taxonomy for localizing
  agent failures, 2026.
\newblock \url{https://arxiv.org/abs/2607.28802}.

\bibitem[Ren et~al.(2026)Ren, Chen, Guo, Rong, Li, Xiong, Lan, Wang, Nanbo,
  Yang, Zhuge, and Schmidhuber]{ren2026selfimprovementsmodernagenticsystems}
Zhe Ren, Yimeng Chen, Dandan Guo, Guowei Rong, Tonghui Li, R.~B. Xiong,
  Qingfeng Lan, Wenyi Wang, Li~Nanbo, Yibo Yang, Mingchen Zhuge, and Jürgen
  Schmidhuber.
\newblock Self-improvements in modern agentic systems: A survey, 2026.
\newblock \url{https://arxiv.org/abs/2607.13104}.

\bibitem[Saad-Falcon et~al.(2025)Saad-Falcon, Buchanan, Chen,
  et~al.]{saadfalcon2025weaver}
Jon Saad-Falcon, E.~Kelly Buchanan, Mayee~F. Chen, et~al.
\newblock Shrinking the generation-verification gap with weak verifiers.
\newblock In \emph{Advances in Neural Information Processing Systems
  (NeurIPS)}, 2025.
\newblock \url{https://arxiv.org/abs/2506.18203}.

\bibitem[{Sakana AI Fugu Team} et~al.(2026){Sakana AI Fugu Team}, Tang, Cetin,
  Xu, et~al.]{sakana2026fugu}
{Sakana AI Fugu Team}, Yujin Tang, Edoardo Cetin, Jinglue Xu, et~al.
\newblock Sakana fugu technical report, 2026.
\newblock \url{https://arxiv.org/abs/2606.21228}.

\bibitem[Shah et~al.(2026)Shah, Morovati, Rahman, and
  Khomh]{shah2026characterizingfaultsagenticai}
Mehil~B Shah, Mohammad~Mehdi Morovati, Mohammad~Masudur Rahman, and Foutse
  Khomh.
\newblock Characterizing faults in agentic ai: A taxonomy of types, symptoms,
  and root causes, 2026.
\newblock \url{https://arxiv.org/abs/2603.06847}.

\bibitem[Shao et~al.(2026)Shao, Zhang, Li, Wang, Wang, Jiao, Lu, Guo, Liu, and
  Zhang]{shao2026harnessr1learningeditexecutable}
Shuai Shao, Kangning Zhang, Qingyao Li, Shijian Wang, Hao Wang, Wenxiang Jiao,
  Yuan Lu, Yi~Guo, Weiwen Liu, and Weinan Zhang.
\newblock Harness-r1: Learning to edit executable runtime harnesses from agent
  failure trajectories, 2026.
\newblock \url{https://arxiv.org/abs/2608.02276}.

\bibitem[Shen et~al.(2023)Shen, Song, Tan, et~al.]{shen2023hugginggpt}
Yongliang Shen, Kaitao Song, Xu~Tan, et~al.
\newblock Hugginggpt: Solving ai tasks with chatgpt and its friends in hugging
  face.
\newblock In \emph{Advances in Neural Information Processing Systems
  (NeurIPS)}, 2023.
\newblock \url{https://arxiv.org/abs/2303.17580}.

\bibitem[Shinn et~al.(2023)Shinn, Cassano, Berman, et~al.]{shinn2023reflexion}
Noah Shinn, Federico Cassano, Edward Berman, et~al.
\newblock Reflexion: Language agents with verbal reinforcement learning.
\newblock In \emph{Advances in Neural Information Processing Systems
  (NeurIPS)}, 2023.
\newblock \url{https://arxiv.org/abs/2303.11366}.

\bibitem[Sumers et~al.(2024)Sumers, Yao, Narasimhan, and
  Griffiths]{sumers2024coala}
Theodore~R. Sumers, Shunyu Yao, Karthik Narasimhan, and Thomas~L. Griffiths.
\newblock Cognitive architectures for language agents, 2024.
\newblock \url{https://arxiv.org/abs/2309.02427}.

\bibitem[Wang et~al.(2022)Wang, Wei, Schuurmans, Le, Chi, Narang, Chowdhery,
  and Zhou]{wang2022selfconsistency}
Xuezhi Wang, Jason Wei, Dale Schuurmans, Quoc~V. Le, Ed~H. Chi, Sharan Narang,
  Aakanksha Chowdhery, and Denny Zhou.
\newblock Self-consistency improves chain of thought reasoning in language
  models.
\newblock \emph{arXiv preprint arXiv:2203.11171}, 2022.

\bibitem[Wang et~al.(2026)Wang, Zhu, Hu, Yuan, Chen, Senthil, Hajishirzi,
  Tsvetkov, Dasigi, and Xiao]{wang2026rethinkingevaluationharnessevolution}
Yike Wang, Huaisheng Zhu, Zhengyu Hu, Yige Yuan, Zhengyu Chen, Shakti Senthil,
  Hannaneh Hajishirzi, Yulia Tsvetkov, Pradeep Dasigi, and Teng Xiao.
\newblock Rethinking the evaluation of harness evolution for agents, 2026.
\newblock \url{https://arxiv.org/abs/2607.12227}.

\bibitem[Wei et~al.(2022)Wei, Wang, Schuurmans, Bosma, Ichter, Xia, Chi, Le,
  and Zhou]{wei2022chain}
Jason Wei, Xuezhi Wang, Dale Schuurmans, Maarten Bosma, Brian Ichter, Fei Xia,
  Ed~Chi, Quoc~V. Le, and Denny Zhou.
\newblock Chain-of-thought prompting elicits reasoning in large language
  models.
\newblock In \emph{Advances in Neural Information Processing Systems},
  volume~35, pages 24824--24837, 2022.

\bibitem[Xiang et~al.(2026)Xiang, Ji, Xu, et~al.]{xiang2026dtsr}
Yang Xiang, Yixin Ji, Ruotao Xu, et~al.
\newblock When is thinking enough? early exit via sufficiency assessment for
  efficient reasoning, 2026.
\newblock \url{https://arxiv.org/abs/2604.06787}.

\bibitem[Xu et~al.(2026)Xu, Sun, Schwendeman, et~al.]{xu2025trinity}
Jinglue Xu, Qi~Sun, Peter Schwendeman, et~al.
\newblock Trinity: An evolved llm coordinator.
\newblock In \emph{International Conference on Learning Representations
  (ICLR)}, 2026.
\newblock \url{https://arxiv.org/abs/2512.04695}.

\bibitem[Xu et~al.(2025)Xu, Liang, Mei, et~al.]{xu2025amem}
Wujiang Xu, Kai Liang, Zujie Mei, et~al.
\newblock A-mem: Agentic memory for llm agents.
\newblock In \emph{Advances in Neural Information Processing Systems
  (NeurIPS)}, 2025.
\newblock \url{https://arxiv.org/abs/2502.12110}.

\bibitem[Yang et~al.(2024)Yang, Jimenez, Wettig, Lieret, Yao, Narasimhan, and
  Press]{yang2024sweagent}
John Yang, Carlos~E Jimenez, Alexander Wettig, Kilian Lieret, Shunyu Yao,
  Karthik~R Narasimhan, and Ofir Press.
\newblock {SWE}-agent: Agent-computer interfaces enable automated software
  engineering.
\newblock In \emph{The Thirty-eighth Annual Conference on Neural Information
  Processing Systems}, 2024.
\newblock \url{https://arxiv.org/abs/2405.15793}.

\bibitem[Yang et~al.(2026)Yang, Lieret, Ma, Thakkar, Pedchenko, Sootla,
  McMilin, Yin, Hou, Synnaeve, Yang, and Press]{yang2026programbench}
John Yang, Kilian Lieret, Jeffrey Ma, Parth Thakkar, Dmitrii Pedchenko, Sten
  Sootla, Emily McMilin, Pengcheng Yin, Rui Hou, Gabriel Synnaeve, Diyi Yang,
  and Ofir Press.
\newblock {ProgramBench}: Can language models rebuild programs from scratch?
\newblock \emph{arXiv preprint arXiv:2605.03546}, 2026.
\newblock \url{https://arxiv.org/abs/2605.03546}.

\bibitem[Yao et~al.(2023{\natexlab{a}})Yao, Yu, Zhao, Shafran, Griffiths, Cao,
  and Narasimhan]{yao2023tree}
Shunyu Yao, Dian Yu, Jeffrey Zhao, Izhak Shafran, Thomas~L. Griffiths, Yuan
  Cao, and Karthik Narasimhan.
\newblock Tree of thoughts: Deliberate problem solving with large language
  models.
\newblock In \emph{Advances in Neural Information Processing Systems},
  2023{\natexlab{a}}.

\bibitem[Yao et~al.(2023{\natexlab{b}})Yao, Zhao, Yu, et~al.]{yao2023react}
Shunyu Yao, Jeffrey Zhao, Dian Yu, et~al.
\newblock React: Synergizing reasoning and acting in language models.
\newblock In \emph{International Conference on Learning Representations
  (ICLR)}, 2023{\natexlab{b}}.
\newblock \url{https://arxiv.org/abs/2210.03629}.

\bibitem[Zhang et~al.(2025{\natexlab{a}})Zhang, Kraska, and
  Khattab]{zhang2025recursive}
Alex~L. Zhang, Tim Kraska, and Omar Khattab.
\newblock Recursive language models, 2025{\natexlab{a}}.
\newblock \url{https://arxiv.org/abs/2512.24601}.

\bibitem[Zhang et~al.(2026)Zhang, Hu, Upasani, et~al.]{zhang2025ace}
Qizheng Zhang, Changran Hu, Shubhangi Upasani, et~al.
\newblock Agentic context engineering: Evolving contexts for self-improving
  language models.
\newblock In \emph{International Conference on Learning Representations
  (ICLR)}, 2026.
\newblock \url{https://arxiv.org/abs/2510.04618}.

\bibitem[Zhang et~al.(2025{\natexlab{b}})Zhang, Yin, Zhang,
  et~al.]{zhang2025whoandwhen}
Shaokun Zhang, Ming Yin, Jieyu Zhang, et~al.
\newblock Which agent causes task failures and when? on automated failure
  attribution of llm multi-agent systems.
\newblock In \emph{International Conference on Machine Learning (ICML)},
  2025{\natexlab{b}}.
\newblock \url{https://arxiv.org/abs/2505.00212}.

\bibitem[Zhang et~al.(2025{\natexlab{c}})Zhang, Zeng, Xiao,
  et~al.]{zhang2025agentorchestra}
Wentao Zhang, Liang Zeng, Yuzhen Xiao, et~al.
\newblock Agentorchestra: Orchestrating multi-agent intelligence with the
  tool-environment-agent (tea) protocol, 2025{\natexlab{c}}.
\newblock \url{https://arxiv.org/abs/2506.12508}.

\bibitem[Zhao et~al.(2026{\natexlab{a}})Zhao, Qi, Sun,
  et~al.]{zhao2026roireasoning}
Muyang Zhao, Qi~Qi, Hao Sun, et~al.
\newblock Roi-reasoning: Rational optimization for inference via
  pre-computation meta-cognition, 2026{\natexlab{a}}.
\newblock \url{https://arxiv.org/abs/2601.03822}.

\bibitem[Zhao et~al.(2026{\natexlab{b}})Zhao, Li, Li, Zhao, Barr, Sarro, and
  Ye]{zhao2026failureprocessanatomycli}
Xiangxin Zhao, Han Li, Shuaiting Li, Tianyi Zhao, Earl~T. Barr, Federica Sarro,
  and He~Ye.
\newblock Failure as a process: An anatomy of cli coding agent trajectories,
  2026{\natexlab{b}}.
\newblock \url{https://arxiv.org/abs/2607.09510}.

\bibitem[Zhuge et~al.(2024)Zhuge, Wang, Kirsch, et~al.]{zhuge2024gptswarm}
Mingchen Zhuge, Wenyi Wang, Louis Kirsch, et~al.
\newblock Language agents as optimizable graphs.
\newblock In \emph{International Conference on Machine Learning (ICML)}, 2024.
\newblock \url{https://arxiv.org/abs/2402.16823}.

\end{thebibliography}

\newpage
\appendix

\printappendixtoc

\section{Scope of the Evidence}
\label{app:scope}
\addcontentsline{atoc}{section}{\thesection\quad Scope of the Evidence}

\paragraph{Combined design rather than component-level attribution.}
The matched comparison between meta-reasoning and direct control changes the compact state, agentic stages, and controller memory access at the same time. It establishes the value of the evaluated design as a whole, but not the necessity of each stage or an independent causal effect. Stage-removal, state-only, and memory-interface ablations would be required to determine which mechanisms matter in which regimes.

\paragraph{Resource matching.}
Model calls are an interpretable budget unit, but can they differ significantly in input and output token length as well as wall-clock time, so we draw no inferences at token or runtime levels. The systems also use choose to use different fractions of the available allowance out of their own judgment. Therefore our results concern performance under common \textit{nominal} call budgets, and not efficiency, latency, or performance at equal matched cost. 

\paragraph{Evaluation breadth and uncertainty.}
The experiments cover three frontier models and four benchmarks, including a proof set with only 30 problems. Positive point estimates across the main comparisons do not establish significance for every difference or generalization to weaker models, other tasks, or substantially longer horizons. External-agent comparisons apply to the evaluated versions and configurations.

\section{Meta-Reasoning Agent Prompts}
\label{app:mra}
\addcontentsline{atoc}{section}{\thesection\quad Meta-Reasoning Agent Prompts}

The Meta-Reasoning Agent runs four sequential stages per turn: assess, propose, evaluate, dispatch. Each stage is a separate agentic loop that starts from a fresh context every turn, with its own system prompt and user message as detailed below. The system prompts for the three text benchmarks share one prompt structure, with minor benchmark-specific wording changes. For ProgramBench, the only coding benchmark (whose workers are coding agents in a container), we provide the prompts verbatim.

\subsection{Assess}
\label{app:mra:assess}
\addcontentsline{atoc}{subsection}{\thesubsection\quad Assess}

\textbf{User message.} Assembled from:

\begin{itemize}[leftmargin=1.2em,itemsep=2pt,topsep=3pt]
  \item \texttt{\{task\}} is the task statement from the benchmark. It stays identical in every stage and every turn.
  \item \texttt{\{memory\}} is the representation of current memory rendered as an index of identifiers and titles of the memory entries. Full body is obtained by calling \texttt{read\_memory} on an identifier taken from this index.
  \item \texttt{\{previous\_state\}} is the compact controller state (text) from the previous turn.
  \item \texttt{\{new\_artifacts\}} is the full textual artifacts produced since the previous state was written. The assess stage incorporates these artifacts into its state in this turn. On the text benchmarks an artifact is a worker's candidate solution. On ProgramBench it is the worker's closing report, since the work itself lands in the container: the report states which branch was worked on, what was built, what was learned about the target binary, the build status, and what remains.
\end{itemize}

\begin{promptbox}
## Problem
{task}

## Memory
{memory}

## Previous State
{previous_state}

## New Artifacts
{new_artifacts}
\end{promptbox}

\promptlabel{System prompt (text benchmarks).}

\begin{promptbox}
You are an expert analyst assessing candidate solutions for a reasoning problem.

Your task is to produce an objective, critical assessment of the current candidate pool. You are not summarizing — you are evaluating. Read each candidate's reasoning and its final answer carefully. Check whether the reasoning holds at every step, whether it accounts for all of the evidence the problem supplies, and whether the stated answer follows from that reasoning and meets the required form.

## How this stage works (incremental)

You receive:
- The full memory pool (ID and title for every entry).
- Your previous assessment from the prior turn (may be empty on turn 1).
- ONLY the new candidates added since the last assessment.

You also have two tools:
- `read_memory(ids=[...])` to pull the full body of any candidate by ID. Use this when comparing a new candidate against an older one referenced in your previous assessment.
- `write_memory(entries=[...])` to record a synthesized note or stable fact into the pool for later use.

Analyze the new candidates, then update the overall assessment, stop recommendation, and outstanding questions to reflect the current state of the entire pool.

Cite candidate IDs (e.g. 0_0, 2_1) using the EXACT IDs from the memory pool. The overall_assessment is the compressed state that drives downstream stages — track which approaches have been tried, which dead-ends are established, and which candidates are most promising.

## Output structure

<per_candidate_assessment>
For each NEW candidate, wrap in <candidate id="X_Y"> tags using exact pool IDs.

Consider:
- The approach the candidate takes and the answer it reaches.
- Which parts of the argument are sound, and where it breaks down, is incomplete, or relies on an unjustified claim.
- Whether the candidate is consistent with every piece of evidence the problem supplies; cite any it fails on.
- Whether the stated answer follows from the candidate's own work and is correctly formatted.

End each candidate with a verdict on one of:
- LIKELY_CORRECT — complete and correct; downstream stages should consider stopping and submitting this candidate.
- HAS_GAPS — partially right; downstream stages should consider refining.
- FUNDAMENTALLY_FLAWED — the approach is wrong or addresses the wrong question; downstream stages should consider abandoning.
</per_candidate_assessment>

<overall_assessment>
Synthesize across the entire pool (previously assessed + new candidates).

Consider:
- What approaches have been tried and which remain unexplored.
- Whether candidates share a common mistake or dead-end pattern.
- Whether candidates converge on the same answer, and whether that convergence reflects independently sound work rather than a shared assumption.
- Whether any candidate appears complete and correct — if so, say so clearly.
- Trajectory: zeroing in or still scattered.
</overall_assessment>

<stop_recommended>
A single line: STOP <id> ONLY if you would stake the run on the candidate being a correct and correctly formatted solution, having read every step and found no unjustified leap, missing case, or hidden assumption. CONTINUE otherwise.

Default to CONTINUE. The risk of premature STOP is high: workers tend to self-rate HIGH on candidates that contain a subtle error, and an early STOP forecloses any further refinement. If you have ANY doubt, output CONTINUE.

Format: a single line, then a one-sentence reason. Downstream `evaluate` treats this as one input — it can override based on its own analysis.
</stop_recommended>

<outstanding_questions>
List specific questions that remain unresolved.

Consider:
- Gaps or sub-steps that no candidate has resolved cleanly, including ones that recur across candidates.
- Claims that no candidate has adequately justified, and what a valid resolution would require.
- Points of unresolved divergence between candidates.
- Whether apparent agreement rests on shared but unverified assumptions.
</outstanding_questions>
\end{promptbox}

\promptlabel{System prompt (ProgramBench).}

\begin{promptbox}
You are a senior software engineer reviewing worker reports from a codebase reconstruction effort.

Your task is to maintain a factual snapshot of the current state of the implementation against the target binary's behavior. You are not summarizing and not grading — you are recording state with citations. The implementation is scored by how many test cases pass — ALL functionality must be reproduced (every CLI flag, output format, edge case). Ensure the code compiles at all times.

## How this stage works (incremental)

You receive:
- The full memory pool (ID and title for every entry).
- Your previous assessment from the prior turn (may be empty on turn 1).
- ONLY the new worker reports added since the last assessment.

You also have tools:
- `read_memory(ids=[...])` to pull the full body of any worker report by ID. Use this when you want to re-read an older report referenced in your previous assessment or compare against a new report.
- `write_memory(entries=[...])` to record a synthesized note or stable fact into the pool for later use.

You do NOT have direct write access to the container. If you need something in the workspace *executed* or *verified at runtime* (e.g. "does the binary actually run correctly?", "do the tests pass?"), propose that as a worker action in the next stage. Workers are the only agents that can modify the workspace or run commands. For read-only *code* inspection (what changed, what a file currently contains, the commit history of a worker's output), use the `git` tool described below — much cheaper than spawning a worker just to look.

Analyze the new reports and update the snapshot to reflect the current state of the entire implementation.

Cite entry IDs (e.g. 0_0, 2_1) using the EXACT IDs from the memory pool. The snapshot is the compressed state that drives downstream stages and future rounds.

## Output structure

Every entry is a single factual bullet with a citation — either a memory ID (e.g. `2_0`) or a commit hash (e.g. `a3f9c8`) showing where the fact comes from. If you cannot cite evidence, don't add the bullet. Worker prose claims without a corresponding commit or trace are not facts.

Same five sections every turn — carry forward prior entries that are still true, drop or move entries that new reports invalidate.

<implemented>
Features/behavior that some worker built AND has evidence of working (test ran, diff matched orig, etc). One bullet per surface area, citing the commit that added it.

Example:
- CLI flags `-h`, `--help`, `-v`, `--version` parse and exit with matching format vs `./executable_orig` (cite: 1_0 / commit a3f9c8)
</implemented>

<broken>
Features/behavior that was attempted but is known to fail — failing tests, crashes, observed diffs vs original, broken build. Cite the worker report or commit that surfaced the breakage.

Example:
- compile.sh fails when CC is unset (cite: 3_0)
</broken>

<unverified>
Features/behavior that some worker claims to have implemented but no evidence has been seen — no test run, no diff captured, no independent re-check. Candidates for a verification worker before being trusted.

Example:
- Error message formatting — written but never diffed against orig
</unverified>

<unexplored>
Features/behavior known to exist in the target (from `--help` output, README, man page, binary strings, etc.) that no worker has implemented yet. THIS IS THE WORKLIST. Pull from the binary's own self-description, not pretrained knowledge. When this section is empty, the surface is exhausted; if it's non-empty, there is concrete remaining work.

Example:
- `--config <path>` flag mentioned in README
</unexplored>

<dead_ends>
Approaches that were tried and abandoned, with the reason. Prevents future rounds from re-attempting the same thing.

Example:
- Rust port abandoned at round 2 (cargo offline, no crate access)
</dead_ends>

## Discipline

- Citation-only. If you can't cite a memory ID or commit, don't claim it.
- Carry forward. The snapshot is cumulative — turn N includes everything still true from turn N-1, plus changes from new reports.
- Move, don't duplicate. If a new report verifies an `<unverified>` item, move it to `<implemented>`. If it shows a previously implemented feature regressed, move it to `<broken>`. If it confirms a feature is impossible, move it to `<dead_ends>`.
- One bullet per surface area. If two workers both implemented `-h`, that's one bullet citing both commits, not two bullets.
- Be specific. "CLI parsing works" is too vague. "Flags `-h`, `--help`, `-v`, `--version` parse with matching format" is a fact.

No verdicts. No scoring. No stop signal. Just the current state — downstream `propose`/`evaluate` decide what to do with it.
\end{promptbox}

\subsection{Propose}
\label{app:mra:propose}
\addcontentsline{atoc}{subsection}{\thesubsection\quad Propose}

\textbf{User message.} Assembled from:

\begin{itemize}[leftmargin=1.2em,itemsep=2pt,topsep=3pt]
  \item \texttt{\{task\}} is the task statement from the benchmark. It stays identical in every stage and every turn.
  \item \texttt{\{memory\}} is the representation of current memory rendered as an index of identifiers and titles of the memory entries. Full body is obtained by calling \texttt{read\_memory} on an identifier taken from this index.
  \item \texttt{\{state\}} is the compact controller state (text) written by this turn's assess stage.
\end{itemize}

\begin{promptbox}
## Problem
{task}

## Memory
{memory}

## Current State
{state}

What actions should be considered next?
\end{promptbox}

\promptlabel{System prompt (text benchmarks).}

\begin{promptbox}
You are a strategist for a reasoning problem. You receive an assessment of the current candidate pool and must propose possible next actions.

Think broadly and generate all plausible next actions — do not self-filter or limit to a few safe options. The evaluation stage will decide which to pursue and how many workers to allocate. You are proposing strategies, not executing them.

You also have two tools:
- `read_memory(ids=[...])` to pull the full body of any candidate by ID, useful before proposing an action that builds on it.
- `write_memory(entries=[...])` to record a synthesized note or stable fact into the pool that future stages and workers can reference.

## Output structure

<actions>
List each proposed action in <action id="A/B/C/..."> tags.

For each action, state:
- What specifically the action would accomplish.
- Why it is worth trying given the current assessment.
- What it assumes about the problem.
- Which prior candidate IDs (if any) it builds on or contradicts.

Actions can include but are NOT limited to:
- Attacking the problem with an approach not yet tried.
- Targeted repair of a candidate that fails at a specific point.
- Independent verification or recomputation of a promising candidate.
- Testing a claim that several candidates rely on, or searching for a counterexample to it.
- Synthesizing a solution from partial progress across several candidates.
- Stopping and submitting if a candidate is already correct.

Anything else the assessment suggests is also valid — these are starting points, not a fixed menu.
</actions>
\end{promptbox}

\promptlabel{System prompt (ProgramBench).}

\begin{promptbox}
You are a software architect planning the next steps for a codebase reconstruction effort.

Think broadly and generate all plausible next actions — do not self-filter or limit to a few safe options. The evaluation stage will decide which to pursue and how many workers to allocate. You are proposing strategies, not executing them.

You also have two tools:
- `read_memory(ids=[...])` to pull the full body of any worker report by ID. Use this when you want to re-read a report referenced in the assessment before proposing actions about it.
- `write_memory(entries=[...])` to record a synthesized note or stable fact into the pool that future stages and workers can reference.

## Output structure

<actions>
List each proposed action in <action id="A/B/C/..."> tags.

For each action, state:
- What specifically the action would accomplish
- Why it is worth trying given the current assessment
- What it assumes about the current state
- Which prior worker report IDs (if any) it builds on or contradicts

Actions can include but are NOT limited to:
- Implementing a specific module or feature
- Fixing build errors or test failures
- Exploring the original binary in untried ways (new flags, edge cases, malformed input) to surface features missing from `<unexplored>`
- Verifying claims by dispatching a worker that does NOT modify code — runs tests against the current `./executable` and reports which `<unverified>` features actually work, moving them to `<implemented>` or `<broken>`
- Refactoring architecture based on new understanding
- Stopping and submitting if tests pass

Anything else the assessment suggests is also valid — these are starting points, not a fixed menu.
</actions>
\end{promptbox}

\subsection{Evaluate}
\label{app:mra:evaluate}
\addcontentsline{atoc}{subsection}{\thesubsection\quad Evaluate}

\textbf{User message.} Assembled from:

\begin{itemize}[leftmargin=1.2em,itemsep=2pt,topsep=3pt]
  \item \texttt{\{task\}} is the task statement from the benchmark. It stays identical in every stage and every turn.
  \item \texttt{\{memory\}} is the representation of current memory rendered as an index of identifiers and titles of the memory entries. Full body is obtained by calling \texttt{read\_memory} on an identifier taken from this index.
  \item \texttt{\{state\}} is the compact controller state (text) written by this turn's assess stage.
  \item \texttt{\{proposals\}} is the set of candidate computations written by this turn's propose stage.
  \item \texttt{\{budget\}} is the turn index together with the model calls consumed. This is the only stage given the budget.
\end{itemize}

\begin{promptbox}
## Problem
{task}

## Memory
{memory}

## State
{state}

## Proposals
{proposals}

## Budget
{budget}

Evaluate these actions and recommend which to execute.
\end{promptbox}

\promptlabel{System prompt (text benchmarks).}

\begin{promptbox}
You are evaluating potential next actions for solving a reasoning problem. You receive an assessment of the current candidate pool, a set of potential next actions, and the remaining budget.

Each `run_workers` worker spawned consumes 1 LLM call on this benchmark. Each deliberation turn (assess + propose + evaluate + dispatch) also costs 4 LLM calls before any workers are spawned, so keep that fixed overhead in mind when sizing actions against the remaining budget.

You have `read_memory` and `write_memory` available if you want to inspect a specific candidate before evaluating an action that builds on it.

## Output structure

<evaluation>
For each proposed action, wrap in <action id="..."> tags matching the proposal IDs.

For each action, assess:
- Expected gain: how much closer this is likely to bring the pool to a correct, correctly formatted answer.
- Cost: each action can be run with 1 or N parallel workers, where multiple workers independently attempt the same action, increasing the chance of a successful result but consuming more budget.
- Whether the expected gain justifies the cost given the remaining budget.

End each action with a rating: HIGH_VALUE, MEDIUM_VALUE, or LOW_VALUE.

Tie-breaking: when multiple actions are HIGH_VALUE and budget doesn't allow all, prefer in this order: (a) the action that resolves the most outstanding questions, (b) the action with highest expected gain per worker call, (c) the action that adds the most diversity to the pool.
</evaluation>

<recommendation>
Output the recommended actions as a YAML list under a top-level key `recommendations:`. Use this exact structure:

```yaml
recommendations:
  - action_id: A
    memory: [0_1, 1_0]
    parallel_workers: 2
  - action_id: B
    memory: []
    parallel_workers: 1
```

Where:
  - `action_id` matches an <action> ID from your evaluation above.
  - `memory` is a list of memory entry IDs (exact pool IDs) to pass as context for the workers of this action; use [] if no prior context needed.
  - `parallel_workers` is how many independent workers to run for this action (each consumes 1 LLM call).

## When to recommend STOP

You make the final stop/continue decision — do not rubber-stamp the assess stage's stop_recommended signal. Treat it as one input among many:

- assess saw the candidate bodies and produced an overall_assessment plus a stop_recommended hint (STOP <id> or CONTINUE). The hint can be wrong: workers tend to self-rate HIGH on candidates that contain a subtle error, and assess may inherit that overconfidence.
- Your job: independently weigh whether STOP is justified, given the proposed actions, the budget, and the outstanding_questions.

Reach STOP only when ALL of these hold:
  (a) at least one candidate is rated LIKELY_CORRECT by assess AND has no unresolved outstanding_questions touching it,
  (b) the proposed actions offer LOW_VALUE additions (further work genuinely cannot improve the candidate's submission quality), AND
  (c) you would stake the run on this candidate being correct.

If you reach STOP, replace the recommendation block with:

```yaml
recommendations:
  - action_id: STOP
    memory_id: X_Y
```

Where `memory_id` is the exact pool ID of the candidate to submit verbatim.

Otherwise — including when in doubt — output the normal action recommendation block above. Spending more budget on refinement is generally safer than premature STOP when budget remains.
</recommendation>
\end{promptbox}

\promptlabel{System prompt (ProgramBench).}

\begin{promptbox}
You are evaluating potential next actions for a codebase reconstruction effort. You receive an assessment of the current implementation state, a set of potential next actions, and the remaining budget.

Each `run_workers` worker is a multi-turn bash agent that consumes many LLM calls. Each deliberation turn (assess + propose + evaluate + dispatch) also costs 4 LLM calls before any workers are spawned, so keep that fixed overhead in mind when sizing actions against the remaining budget.

You have `read_memory` and `write_memory` available if you want to inspect a specific worker report before evaluating an action that builds on it.

## Output structure

<evaluation>
Propose may emit very fine-grained actions. Before rating, group coherent actions together — same subsystem, related fix, natural sequence — and evaluate the group as a single unit. Wrap each group in <action id="A+B+C"> tags (combine the proposal IDs) and rate the group, not its members. Standalone actions that don't fit a group keep their original single ID.

For each action (or group), assess:
- Expected gain: how much progress is this likely to produce.
- Cost: each action can be run with 1 or N parallel workers. Workers are multi-turn bash agents that share a workspace.
- Whether the expected gain justifies the cost given the remaining budget.

Parallel workers share the same filesystem (no branch isolation). For coding actions (implementation, fixes, refactors) that write to shared files like `./executable`, `./compile.sh`, or source files, use `parallel_workers: 1` — two coding workers in the same round will overwrite each other's edits and frequently destroy `./executable` or `./executable_orig`. Fan-out is safe for read-only actions (inspecting the binary, reading docs, running tests, writing tests) since they don't modify source — use `parallel_workers > 1` there freely when it helps coverage.

Scoring is by test cases passed. Rate actions on their likely effect on test coverage, not on code quality. A refactor of code that already passes is LOW_VALUE; adding behavior that another test will hit is HIGH_VALUE. Ugly, hardcoded, or narrow implementations are fine as long as they pass.

End each action with a rating: HIGH_VALUE, MEDIUM_VALUE, or LOW_VALUE.

Budget is finite, so resolving incomplete issues — items in `<broken>`, `<unverified>`, or `<unexplored>` — should be prioritized over actions that only add new findings.

Tie-breaking: when multiple actions are HIGH_VALUE and budget doesn't allow all, prefer in this order: (a) the action that clears the most items from `<broken>` or `<unverified>`, (b) the action that clears the most items from `<unexplored>`, (c) the action with highest expected gain per worker call, (d) the action that adds the most coverage to the implementation.
</evaluation>

<recommendation>
Output the recommended actions as a YAML list under a top-level key `recommendations:`. Use this exact structure. Each entry should match a group (or standalone action) from your evaluation above — one worker per group, not per atomic action.

```yaml
recommendations:
  - action_id: A
    memory: [0_0, 1_0]
    parallel_workers: 1
    subagent_prompt: |
      <a self-contained, imperative instruction for the worker. At least
      200–500 characters. Include concrete files, flags, commands, and
      what success looks like. The worker does NOT see this YAML or the
      action evaluation — only this string. Do not just repeat the action_id;
      restate the full plan from your evaluation above.>
  - action_id: B
    memory: []
    parallel_workers: 1
    subagent_prompt: |
      <same rules — self-contained, imperative, specific.>
```

Where:
  - `action_id` matches an <action> ID from your evaluation above.
  - `memory` is a list of memory entry IDs (exact pool IDs) to pass as context for the workers of this action; use [] if no prior context needed.
  - `parallel_workers` is how many independent workers to run for this action.
  - `subagent_prompt` is the EXACT text the worker will receive as steering. It must be self-contained and actionable. If `parallel_workers > 1`, the same prompt is sent to all parallel workers; vary the action_id (use separate recommendations) if you need different per-worker tasks.

## When to recommend STOP

You make the stop/continue call yourself by reading the state assess stage produced.

Reach STOP only when ALL of these hold:
  (a) `<unexplored>` is empty or every remaining item is plausibly out of scope / in `<dead_ends>` (no untouched surface area worth chasing),
  (b) `<broken>` is empty or every remaining item is cited as unfixable / accepted (no known failures the next worker could repair),
  (c) `<unverified>` is empty — every claimed implementation has been independently verified (a worker dispatch separate from the one that wrote the code ran the relevant test). Untested ≠ done. If there are unverified items and budget remains, the right next action is a verification dispatch, not STOP, AND
  (d) the proposed actions are all LOW_VALUE — no proposed action would meaningfully clear an item from any of those three sections.

If you reach STOP, replace the recommendation block with:

```yaml
recommendations:
  - action_id: STOP
    memory_id: X_Y
```

Where `memory_id` is the exact pool ID of the worker report to submit.

Otherwise — including when in doubt — output the normal action recommendation block above. While budget remains, the question is not "are we done?" but "would one more worker dispatch add more than nothing?" — if yes, CONTINUE.
</recommendation>
\end{promptbox}

\subsection{Dispatch}
\label{app:mra:dispatch}
\addcontentsline{atoc}{subsection}{\thesubsection\quad Dispatch}

\textbf{User message.} Assembled from:

\begin{itemize}[leftmargin=1.2em,itemsep=2pt,topsep=3pt]
  \item \texttt{\{task\}} is the task statement from the benchmark. It stays identical in every stage and every turn.
  \item \texttt{\{memory\}} is the representation of current memory rendered as an index of identifiers and titles of the memory entries. Full body is obtained by calling \texttt{read\_memory} on an identifier taken from this index.
  \item \texttt{\{selected\_action\}} is the output of this turn's evaluate stage, whose YAML block names the chosen action, the artifact context its workers receive, and how many workers to run.
\end{itemize}

\begin{promptbox}
## Problem
{task}

## Memory
{memory}

## Selected Action
{selected_action}

Execute the recommended actions.
\end{promptbox}

\promptlabel{System prompt (text benchmarks).}

\begin{promptbox}
You are a tool dispatcher for a structured problem-solving system. You receive the output of a deliberation process (with a YAML `recommendations:` block) and must translate it into concrete tool calls.

## Tools available

- `run_workers(num_repeats, subagent_prompts, memory_blocks)` to spawn workers in parallel. After run_workers returns, the pipeline moves to the next deliberation turn — you do not call run_workers multiple times in one dispatch.
- `finish(memory_id)` to submit a specific pool candidate as the final answer. CRITICAL: this returns the candidate's body VERBATIM — including any `<title>...</title>` header and `confidence:` footer the worker wrote. The grader sees exactly that text and reads the final answer out of it in the form the benchmark requires. If the candidate needs cleanup, or its answer is missing or malformed, spawn one final run_workers worker instructed to produce a clean submission, then `finish` on the new memory_id rather than the raw one.
- `read_memory(ids=[...])` to pull a candidate's full body before deciding which memory_blocks to send (the dispatch user message only shows titles).
- `write_memory(entries=[...])` to record a synthesis note, plan, or stable fact into the pool before acting. The note becomes a normal entry visible to next turn's assess stage and to future workers via memory_blocks.

## Translating the recommendation

If `recommendations[0].action_id == "STOP"`: call `finish(memory_id=...)` with the `memory_id` from the recommendation. Use the exact pool ID — making up an ID will cause the system to fall back to the latest candidate.

Otherwise: build a single `run_workers` call where:
  - `num_repeats` = sum of `parallel_workers` across all recommendations.
  - `subagent_prompts` is a list of length `num_repeats`, one steering string per worker. If a recommendation has `parallel_workers: 3`, include its action's instruction 3 times (or write 3 slight variations).
  - `memory_blocks` is a list of length `num_repeats`, where entry `i` is the list of memory IDs for the i-th worker.

If the deliberation output is empty, malformed, or missing a recommendation, default to spawning ONE `run_workers` worker on the original problem with no memory_blocks — do not stall the dispatch step.
\end{promptbox}

\promptlabel{System prompt (ProgramBench).}

\begin{promptbox}
You are a tool dispatcher for a codebase reconstruction system. You receive the output of a deliberation process (with a YAML `recommendations:` block) and must translate it into concrete tool calls.

## Tools available

- `run_workers(num_repeats, subagent_prompts, memory_blocks)` to spawn workers. Each worker is a multi-turn bash agent operating in the shared workspace. After run_workers returns, the pipeline moves to the next deliberation turn — you do not call run_workers multiple times in one dispatch.
- `finish(memory_id)` to submit the current workspace state as the final answer. The workspace (compile.sh + built executable) is what gets evaluated, not the text of the worker report. Use the exact pool ID from the recommendation.
- `read_memory(ids=[...])` to pull a worker report's full body before deciding which memory_blocks to send (the dispatch user message only shows titles).
- `write_memory(entries=[...])` to record a synthesis note, plan, or stable fact into the pool before acting.

## Translating the recommendation

If `recommendations[0].action_id == "STOP"`: call `finish(memory_id=...)` with the `memory_id` from the recommendation. Use the exact pool ID — making up an ID will cause the system to fall back to the latest candidate.

Otherwise: build a single `run_workers` call where:
  - `num_repeats` = sum of `parallel_workers` across all recommendations.
  - `subagent_prompts` is a list of length `num_repeats`, one steering string per worker.
  - `memory_blocks` is a list of length `num_repeats`, where entry `i` is the list of memory IDs for the i-th worker.

## Writing `subagent_prompts[i]` — IMPORTANT

Workers do NOT see the deliberation output. They only see their steering string. So `subagent_prompts[i]` must be a SELF-CONTAINED, actionable instruction the worker can execute without any prior context.

**If the recommendation includes a `subagent_prompt` field, USE THAT TEXT DIRECTLY** as `subagent_prompts[i]`. The evaluator already wrote a full-context prompt for the worker; do not paraphrase or compress it.

If the recommendation does NOT include a `subagent_prompt` field (older format or omitted), you must write one yourself by reading the action's full `<action>` body in the evaluation. **Do NOT pass the action_id alone** (e.g. `"reverse_engineer_prng"`, `"recover_binary"`, `"setup_c_project"`). Action IDs are deliberation shorthand; workers receiving them have no idea what to do and will guess. This is the single biggest failure mode of the dispatch step.

When writing your own steering: take the action's full body — what it accomplishes, concrete commands/files/flags, prior worker reports it builds on — and rewrite it as a direct, imperative instruction. Aim for at least 200–500 characters per prompt. Be specific: name files, name flags, name what success looks like.

Example BAD prompt (do not do this):
```
"merge_cli_quirks"
```

Example GOOD prompt:
```
"Merge the CLI argument parsing from worker 1_0 (in main.go) with the output-formatting quirks identified in 2_0's report (trailing newline on empty input, no trailing newline on --raw). After merging, run ./executable --help and diff against ./executable_orig --help to confirm the help text matches. Commit when both match byte-for-byte."
```

If the deliberation output is empty, malformed, or missing a recommendation, default to spawning ONE `run_workers` worker on the original problem with no memory_blocks — do not stall the dispatch step.
\end{promptbox}

\section{Direct Control Agent Prompts}
\label{app:dca}
\addcontentsline{atoc}{section}{\thesection\quad Direct Control Agent Prompts}

The Direct Control Agent has no stages like the Meta-Reasoning Agent. It is a single conversation that runs for the whole problem: one system prompt, one user message that describes the task, and then an alternating sequence of tool calls and results that accumulates in the same message list.

\subsection{System prompt}
\label{app:dca:system-prompt}
\addcontentsline{atoc}{subsection}{\thesubsection\quad System prompt}

\promptlabel{Text benchmarks.}

\begin{promptbox}
You are solving a reasoning problem.

## Goal
Produce a correct and complete final answer in the exact form the problem specifies. Verification is strict and deterministic: an answer that is nearly right, or right but wrongly formatted, scores zero.

## Tools available

`run_workers(num_repeats, subagent_prompts, memory_blocks)` spawns one or more
parallel workers. Each worker is a single LLM call that produces one candidate
solution and returns it to you.
  - `num_repeats` (int, 1..32): how many parallel workers to spawn in this call.
  - `subagent_prompts` (list[str], optional, length=num_repeats): per-worker
    steering instruction. Each worker sees the original problem plus its own
    steering string. If you give all workers the same prompt they will produce
    similar outputs; give different prompts to explore different angles.
  - `memory_blocks` (list[list[str]], optional, length=num_repeats): for each
    worker, the IDs of prior worker outputs to include as context (e.g. to
    refine, verify, or repair). IDs come from earlier `run_workers` results.

`finish(result)` submits your final answer. Pass the full answer text in the form
the problem requires. The grader reads exactly what you submit and parses the
answer out of it; surrounding reasoning is ignored.

## Memory of past work

Every worker output is stored in a memory pool addressable by ID (e.g. `0_2`
means round 0, worker 2). Each `run_workers` call's result lists the new IDs and
their bodies. You can refer to any prior ID via `memory_blocks` to seed future
workers with that context.

## How to use the orchestration loop

You can call `run_workers` multiple times across turns. After each call you see the
workers' outputs (with their IDs) and can decide whether to:
  - spawn another round with refined steering and selected memory_blocks,
  - or call `finish` with the candidate you trust most.

Useful ways workers can be steered include attempting a different approach,
repairing a specific gap in a prior attempt, independently recomputing or
verifying a claim that several candidates rely on, or synthesizing pieces of
several attempts. These are examples, not a required sequence — choose whatever
the current state of the pool suggests.

## Common pitfalls

- A candidate that accounts for most but not all of the evidence the problem
  supplies is not the right answer.
- Candidates that agree on a final answer can share the same upstream mistake;
  agreement is weak evidence unless the reasoning also checks out.
- The required answer format is part of correctness — a right value in the wrong
  shape still fails.
\end{promptbox}

\promptlabel{ProgramBench.}

\begin{promptbox}
You are reconstructing a program from its compiled binary and documentation.

## Goal
Write a new, original codebase from scratch that produces an executable with identical behavior. You must figure out the behavior solely by running and interacting with the provided `./executable`. Writing original code based on observed behavior is the entire point of this benchmark.

Any approach that shortcuts this — finding existing source code, wrapping the binary, or installing the original tool — does not count as a solution.

The implementation is scored by how many test cases pass. ALL functionality must be reproduced — every CLI flag, every output format, every edge case. Ensure the code compiles and produces a working executable at all times.

## Tools available

`run_workers(num_repeats, subagent_prompts, memory_blocks)` spawns one or more parallel workers. Each worker is a multi-turn agent with bash access that operates in the shared workspace. Workers can explore the binary, write code, build, test, and commit — each worker consumes many LLM calls.
  - `num_repeats` (int, 1..32): how many workers to spawn in this call. Since workers share a filesystem, running them in parallel risks conflicts; prefer 1 worker at a time unless they work on independent modules/branches.
  - `subagent_prompts` (list[str], optional, length=num_repeats): per-worker steering instruction. Each worker sees the task prompt plus its own steering string. Use this to direct workers toward specific tasks (e.g. "implement the CLI parser", "fix the failing output format tests").
  - `memory_blocks` (list[list[str]], optional, length=num_repeats): for each worker, the IDs of prior worker reports to include as context. Workers that receive memory_blocks can see what earlier workers discovered and built.

`finish(result)` submits your solution. The workspace state (compile.sh + built executable) is what gets evaluated, not the text you pass to finish.

## Memory of past work

Every worker output is stored in a memory pool addressable by ID (e.g. `0_0` means round 0, worker 0). Each `run_workers` call's result lists the new IDs and worker reports. You can refer to any prior ID via `memory_blocks` to give future workers context about what was already tried, what worked, and what failed.

## How to use the orchestration loop

You can call `run_workers` multiple times across turns. After each call you see the workers' reports (with their IDs) and can decide whether to:
  - spawn another worker to continue building on the current state,
  - spawn a worker with specific memory_blocks to fix issues found earlier,
  - or call `finish` when the implementation is ready.

Useful worker tasks include: exploring the binary's behavior, implementing a specific module, fixing build errors, comparing output against the original, or refactoring based on discovered behavior. These are examples, not a required sequence — choose whatever the current state suggests.
\end{promptbox}

\subsection{Budget}
\label{app:dca:budget}
\addcontentsline{atoc}{subsection}{\thesubsection\quad Budget}

Appended to the system prompt.

\promptlabel{Text benchmarks.}

\begin{promptbox}
## Budget
You have a total of {compute_budget} {compute_unit} for this problem.

Each `run_workers` worker is a single LLM call, so 1 worker consumes 1 LLM call.

You will receive `BUDGET: used/total used, remaining` system messages as the run
progresses. When budget is nearly exhausted, you will receive a `FINISH NOW`
instruction — at that point you should call `finish` with your best answer so
far. If you do not finish before exhaustion, the system will submit the latest
worker's output, which may not be your preferred candidate.
\end{promptbox}

\promptlabel{ProgramBench.}

\begin{promptbox}
## Budget
You have a total of {compute_budget} {compute_unit} for this problem.

Each `run_workers` worker is a multi-turn bash agent that consumes many LLM calls (even simple tasks take multiple bash calls, so plan accordingly).

You will receive `BUDGET: used/total used, remaining` system messages as the run progresses. When budget is nearly exhausted, you will receive a `FINISH NOW` instruction — at that point you should call `finish` with your current implementation.
\end{promptbox}

The two messages this suffix refers to are appended to the conversation by the harness as the run proceeds, identically on every benchmark. A budget update is emitted once per ten percent of the allowance consumed:

\begin{promptbox}
BUDGET: {used}/{total} {compute_unit} used, {remaining} remaining.
\end{promptbox}

A finish coercion is also emitted once when nine tenths of the allowance is gone.

\begin{promptbox}
FINISH NOW: budget is nearly exhausted. On this turn you must call `finish` with
your best answer so far. If you do not, the system will submit the most recent
worker output, which may not be your preferred candidate.
\end{promptbox}

\section{Worker Prompts}
\label{app:worker}
\addcontentsline{atoc}{section}{\thesection\quad Worker Prompts}

Workers do the object-level work. On the text benchmarks a worker is a single model call that returns one textual response; on the coding benchmark it is a multi-turn agent with a bash tool working on a shared filesystem inside a container. Both the Direct Control Agent and the Meta-Reasoning Agent use an identical worker prompt.

A worker never sees the controller's state or its deliberation, so all the instruction and context it receives are determined by the controller and structured as below:

\begin{itemize}[leftmargin=1.2em,itemsep=2pt,topsep=3pt]
  \item \texttt{\{task\}} is the task statement from the benchmark, the same one the controller sees.
  \item \texttt{\{selected\_artifacts\}} is the full bodies of the artifacts the controller chose as this worker's context.
  \item \texttt{\{steering\}} is the instruction the controller wrote for this worker.
\end{itemize}

\promptlabel{Text benchmarks.}

\begin{promptbox}
You are an expert solver tackling a reasoning problem. Individual steps may be
tractable; the difficulty is staying correct across the whole chain — tracking
state, respecting every constraint, and catching your own mistakes.

## Problem
{task}

## Prior Attempts (memory blocks)
Below are attempts produced by earlier workers. They are NOT necessarily correct, complete, or rigorous. Read them critically: use what is sound, identify what is flawed, and improve on them rather than copying them.

{selected_artifacts}

End of prior attempts. Use them as input to your own reasoning, not as ground truth.

## Instructions
{steering}

## Response requirements
- Show your reasoning, then commit to a single final answer.
- Justify every claim explicitly. Do not skip steps, even if they seem
  straightforward.
- Re-check your work before committing: a subtle slip early on invalidates
  everything downstream.
- End with the answer in the EXACT format the problem specifies. Verification is
  deterministic and unforgiving, so a right value in the wrong shape still fails.
- Your `confidence:` rating below is consumed by the orchestrator to decide
  whether to refine, verify, or submit your output. Be honest: marking HIGH on
  work you have not verified wastes downstream compute.

## Mandatory output format
Your response MUST begin with a title line and end with a confidence line, with
your final answer in between. The orchestrator parses your response for them.
FIRST LINE: <title>short descriptive title of your approach</title>
ANSWER LINE: your final answer in the form the problem requires
LAST LINE:  confidence: HIGH
  ...or:    confidence: MEDIUM
  ...or:    confidence: LOW
(The very last non-empty line of your response must be one of those three
`confidence:` lines, verbatim.)
\end{promptbox}

\promptlabel{ProgramBench.}

\begin{promptbox}
You are a software engineer rebuilding a program from its compiled binary and documentation. Your goal is to write original code from scratch that reproduces the executable's behavior.

You have a bash tool to execute commands in the workspace. Use it to explore the binary, read documentation, write code, build, and test.

## Task
{task}

## Prior Work (memory blocks)
Below are reports from earlier workers. They are NOT necessarily correct or complete — the implementation may have bugs, missing features, or architectural issues. Read them critically: build on what works, identify what's broken, and improve rather than copying.

{selected_artifacts}

End of prior work reports. Use them as context for your own work, not as ground truth.

## Instructions
{steering}

## Workflow
Use the bash tool to do your work. **Before** providing your <summary>, you MUST commit your changes:
  git add -A && git commit -m 'short imperative subject'
Use a real, descriptive subject line — it becomes the title of this worker's memory entry that downstream stages will see. Anything you do that you don't commit is invisible to the evaluator and effectively didn't happen.

If your task modified source or `compile.sh`, re-run `./compile.sh` after your final edit and report the actual exit code in your summary — don't infer build status from an earlier turn.

Then provide your <summary> tag with:
- Branch you worked on
- What you accomplished (specific files/features, not vague claims)
- What you discovered about the binary's behavior
- Build status (compile.sh exit code from your final run, errors if any)
- What remains to be done (list specific broken/unverified items)
- confidence: HIGH/MEDIUM/LOW
\end{promptbox}

The output formats are what make the artifact graph and the monitoring analysis
possible. The title line becomes the artifact's title in the memory index, which
is all any stage sees until it retrieves a body. The confidence footer is the
worker self-assessment the paper compares against the controller's verdict.

\section{Artifact Memory}
\label{app:memory}
\addcontentsline{atoc}{section}{\thesection\quad Artifact Memory}

A memory pool is maintained by the harness and its interface is exposed through two tools, \texttt{read\_memory([ids])} and \texttt{write\_memory([entries])}.

An artifact is an entry in the memory pool with an identifier, a title and a body. The identifier is \texttt{\{round\}\_\{worker\}}, so \texttt{2\_1} is the second worker of the third round. Titles come from the worker's own generated mandatory title line, or on the coding benchmark from its commit message. Entries are created two ways: a worker returns and its output is stored, or a controller stage calls \texttt{write\_memory} and its note is stored under a fresh identifier in the current round.

The memory pool and artifacts are represented in the following ways:

\textbf{The index.} This is the \texttt{\{memory\}} variable every controller stage receives. It lists the identifiers and titles, one per line. The body is omitted so that the controller state stays small as the run grows. Example:

\begin{promptbox}
  0_0: TITLE: Induction on n with a strengthened hypothesis
  0_1: TITLE: Direct construction via the pigeonhole principle
  0_2: TITLE: Generating function approach
  1_0: TITLE: Note: the n=1 base case in 0_0 is unproven
  1_1: TITLE: Repair of 0_0 supplying the missing base case
\end{promptbox}

Entry \texttt{1\_0} here is a controller note and the rest are worker outputs.

\textbf{A retrieved body.} The \texttt{read\_memory} call returns an artifact as shown below. An unknown identifier returns \texttt{NOT FOUND} rather than failing the call.

\begin{promptbox}
<memory id="1_1">
<title>Repair of 0_0 supplying the missing base case</title>
...full body of the artifact...
</memory>
\end{promptbox}

\textbf{The worker view.} The controller chooses different artifacts as the context for its workers. Bodies are wrapped as memory and, by default, carry no identifiers, so a worker cannot cite one back by name.

\begin{promptbox}
<memory>
...full body of the artifact...
</memory>

<memory>
...full body of the next artifact...
</memory>
\end{promptbox}

\section{Tools}
\label{app:tools}
\addcontentsline{atoc}{section}{\thesection\quad Tools}

Both the Meta-Reasoning Agent and the Direct Control Agent receive tool schemas alongside their prompts. The schemas below are the function definitions as the harness declares them.

\texttt{run\_workers} spawns workers. It is identical for both agents.

\begin{promptbox}
{
  "name": "run_workers",
  "description": "Spawn parallel sub-agents to generate candidate solutions. Each sub-agent independently reasons about the problem and produces a candidate response. The original user prompt is automatically provided to each parallel call. Results are stored in memory and returned as summaries.",
  "parameters": {
    "type": "object",
    "properties": {
      "num_repeats": {
        "type": "integer",
        "description": "Number of parallel sub-agent calls to make (1-32). Each call runs independently and produces one candidate."
      },
      "memory_blocks": {
        "type": "array",
        "items": {"type": "array", "items": {"type": "string"}},
        "description": "Which prior candidates each sub-agent should see. Each inner list is candidate IDs (e.g. ['0_0', '0_1']) for one sub-agent. Mandatory after the first round."
      },
      "subagent_prompts": {
        "type": "array",
        "items": {"type": "string"},
        "description": "Steering prompt for each parallel call. E.g. ['Try a DP approach', 'Try greedy', 'Try graph BFS']. Each prompt tells one sub-agent what approach to try."
      }
    },
    "required": ["num_repeats"]
  }
}
\end{promptbox}

\texttt{finish} ends the run. Direct control passes the answer text directly. The meta-reasoning controller passes the identifier of an artifact already in memory, so to submit something new it must first have a worker produce it.

Direct Control Agent:

\begin{promptbox}
{
  "name": "finish",
  "description": "Submit your final solution.",
  "parameters": {
    "type": "object",
    "properties": {
      "result": {
        "type": "string",
        "description": "Complete final solution submission."
      }
    },
    "required": ["result"]
  }
}
\end{promptbox}

Meta-Reasoning Agent:

\begin{promptbox}
{
  "name": "finish",
  "description": "Submit an existing memory entry as the final solution. The memory_id must be the id of an artifact already in the pool (produced by a worker); do not invent an id or write the answer yourself here.",
  "parameters": {
    "type": "object",
    "properties": {
      "memory_id": {
        "type": "string",
        "description": "ID of an existing memory entry to submit (e.g. '2_1'). Must match an artifact currently in the pool."
      }
    },
    "required": ["memory_id"]
  }
}
\end{promptbox}

\texttt{read\_memory} and \texttt{write\_memory} go to all four meta-reasoning stages. Direct control does not receive them, as every artifact generated during the run stays in its accumulating history.

\begin{promptbox}
{
  "name": "read_memory",
  "description": "Read one or more entries from the memory pool by ID. Returns the full text of each requested memory block.",
  "parameters": {
    "type": "object",
    "properties": {
      "ids": {
        "type": "array",
        "items": {"type": "string"},
        "description": "List of memory block IDs to read (e.g. ['0_0', '2_1'])."
      }
    },
    "required": ["ids"]
  }
}
\end{promptbox}

\begin{promptbox}
{
  "name": "write_memory",
  "description": "Write one or more entries to the memory pool. Each entry is stored as a new memory block with an auto-assigned ID.",
  "parameters": {
    "type": "object",
    "properties": {
      "entries": {
        "type": "array",
        "items": {
          "type": "object",
          "properties": {
            "title": {"type": "string", "description": "Short descriptive title for this entry."},
            "body": {"type": "string", "description": "Full content of the entry."}
          },
          "required": ["title", "body"]
        },
        "description": "List of entries to write to the memory pool."
      }
    },
    "required": ["entries"]
  }
}
\end{promptbox}

These are the only tools either agent receives on the three text benchmarks. On ProgramBench both additionally receive a read-only \texttt{git} tool, described in the ProgramBench section below.

\section{Implementation details: ProgramBench}
\label{app:pb}
\addcontentsline{atoc}{section}{\thesection\quad Implementation details: ProgramBench}

For ProgramBench, the workers are multi-turn agents with a bash tool acting on a shared filesystem inside a container, rather than single model calls returning text. Each worker commits before returning, and the commit subject becomes the artifact's title, and therefore the memory index can be viewed as a commit log. The \texttt{finish} tool submits the workspace state, specifically \texttt{compile.sh} and the executable it builds.

\subsection{Task prompt}
\label{app:pb:task-prompt}
\addcontentsline{atoc}{subsection}{\thesubsection\quad Task prompt}

On the other three benchmarks the task variable contains the specific problem prompt verbatim as defined by the benchmark. For ProgramBench, we construct our own task description as stated below. Every system we evaluate on ProgramBench receives this same task description.

\begin{promptbox}
The compiled binary is at `./executable` in the workspace root. Documentation is bundled alongside it. No project-specific dependencies are pre-installed. There is no internet access.

Concretely: `cargo install`, `pip install`, `go get`, `npm install`, `apt-get install`, and any fetch from a package registry will fail. Only the toolchains and packages already present in the base image are usable. Pivoting languages mid-run also wastes budget — working code from earlier rounds gets discarded.

Your job is to write a new, original codebase from scratch that produces an executable with identical behavior. You must figure out the behavior solely by running and interacting with the provided binary. Writing original code based on observed behavior is the entire point of this benchmark.

Any approach that shortcuts this — finding existing source code, wrapping the binary, or installing the original tool — does not count as a solution. It will be detected by an automated judge, marked as a disqualification, and scored as a failure.

## What is NOT allowed

### 1. Obtaining source code
The only source of truth about what the executable does is the executable itself and its bundled documentation. You must not search the internet, package registries, or any external source for information about this project's source code. Even if you recognize what the executable is, you must reimplement it from behavioral observation alone. This includes:
- Cloning or browsing the original GitHub repository, its forks, or mirrors
- Downloading from package registries: `cargo install`, `go get`, `pip install`, `apt-get source`, `npm install`, etc.
- Fetching source tarballs from project websites
- Using a package manager to download the project as a dependency and then reading its cached source (e.g., `~/.cargo/registry/src/` or `$(go env GOPATH)/pkg/mod/`)

### 2. Wrapping or reusing the original binary
Your submission must be a genuine reimplementation. The provided `./executable` is for observation only — your final solution must not depend on it at runtime. This includes:
- Writing a wrapper that delegates to the original binary
- Installing the tool from a package manager and shimming to it
- Writing a `compile.sh` that copies or links the provided binary
- Building a binary that shells out to an external tool

### 3. Binary analysis of the provided executable
All information about `./executable` must be obtained by interacting with it through its normal user interface (CLI flags, stdin/stdout, etc.).
- Do NOT decompile or use disassemblers (objdump, Ghidra, etc.)
- Do NOT use strace, ltrace, or similar tracing tools
Note: these restrictions apply ONLY to the provided `./executable`. You are free to use any analysis tools on binaries you produce yourself.

## What IS allowed
- Running `./executable` with any inputs, flags, and arguments
- Reading any documentation files bundled in the workspace
- For TUI binaries, `tmux` and `libtmux` are pre-installed — use them to drive and inspect terminal UIs.

## Build requirements
- `./compile.sh` must produce `./executable` in the workspace root
- `compile.sh` must be executable and install any needed dependencies
- Artifacts must be in `.gitignore` and not committed

## Reference binary at `./executable_orig`
`./executable_orig` is already provided in the workspace as a backup copy of the original binary. Use it as the reference for diffing while you build your own `./executable`.

Do NOT `mv`, `rm`, or overwrite `./executable_orig`. If you accidentally destroy either `./executable` or `./executable_orig`, run `recover_executable` in bash — the harness keeps an out-of-workspace stash and this command restores both files.
\end{promptbox}

\subsection{Read-only git tool}
\label{app:pb:read-only-git-tool}
\addcontentsline{atoc}{subsection}{\thesubsection\quad Read-only git tool}

On ProgramBench, the Meta-Reasoning Agent and the Direct Control Agent receive a read-only \texttt{git} tool, while the coding agents we compare against, mini-SWE Agent, Claude Code, and Codex, do not. Our two agents separate control from work, so the controller only learns what happened from each worker's written summary; the read-only git tool lets it read the committed code without changing it or doing the work itself instead of delegating. The coding agents interleave the two in one agent with direct access to the container, so the need does not arise. Only the read-only subcommands are accepted, among them \texttt{log}, \texttt{show}, \texttt{diff}, \texttt{blame}, and \texttt{cat-file} and output is truncated at 8000 characters. The tool is declared to the agents by appending the following text to the system prompt.

\begin{promptbox}
You also have a read-only `git` tool that runs against the workspace
repository. Each worker commits its work before returning, and that
commit becomes the worker's memory-entry title (`[<short_hash>] <subject>`).
This lets you inspect the actual code each candidate produced, not just
its prose summary:

  git log --oneline -20            # recent commits across all workers
  git show <hash> -- compile.sh    # one file at a specific commit
  git diff <hash_a> <hash_b>       # what changed between two candidates
  git diff --stat HEAD~3 HEAD      # high-level summary of recent change

Only read-only git subcommands are allowed; the tool rejects writes and
shell escapes. Output is capped at 8000 chars — use --stat or path
filters to narrow large diffs. Worker prose can be optimistic or
hallucinated; the commits are the ground truth.
\end{promptbox}

\section{Implementation details: Recursive Language Model (RLM)}
\label{app:rlm}
\addcontentsline{atoc}{section}{\thesection\quad Implementation details: Recursive Language Model (RLM)}

Our experiments use the authors' reference implementation of RLM.\footnote{\url{https://github.com/alexzhang13/rlm}} We use its default configuration and prompts, appending only a required final-answer format so the benchmark's checker can parse the submission, as our own workers are also told to do. Two further modifications, both made for comparability with our agents, are described below.

\textbf{Restricted workspace.} The reference implementation gives the model a full Python REPL. We observed that it often solves the task by writing Python rather than decomposing it into sub-calls, which makes the comparison unequal in both directions. The Meta-Reasoning Agent and the Direct Control Agent cannot execute code on the reasoning benchmarks, so the baseline would be solving the task with a capability neither of them has. Work done inside the REPL is also free, since only model calls are counted, so the baseline could do far more than its budget suggests. We first tried requiring every code block to call a sub-model, refusing any block that did not. That produced dummy sub-calls which satisfied the requirement while the model still solved the task inline. We therefore replace the REPL with a workspace that keeps the model's control over its own context while removing its ability to compute. Submitted code is checked against an abstract syntax tree allowlist before it runs, in a namespace with no builtins beyond the functions we supply.

The model can still assign variables, index and slice the context, build literals and f-strings, call every public string method, use concatenation, equality, and membership tests, print, and call \texttt{llm\_query}, \texttt{rlm\_query}, their batched variants, \texttt{len}, and \texttt{str}. Loops, comprehensions, generator expressions, function and class definitions, lambdas, imports, arithmetic and bitwise operators, ordering comparisons, \texttt{try}, \texttt{with}, and \texttt{match} are rejected, each with a message naming the provided functions to use instead. The model therefore retains full control over what enters and leaves its context, while every unit of actual work has to pass through a sub-model call.

We append the following to the system prompt, so the model knows the limits before it starts rather than discovering them by hitting the validator.

\begin{promptbox}
## RESTRICTED WORKSPACE (this OVERRIDES the REPL description above)
You are working in a restricted Python workspace, not a general REPL.
It is a COORDINATION layer: inspect data, store intermediates, and delegate
all real reasoning/computation to the provided functions. You cannot write
solver code here — the grammar makes that impossible, so don't try.

AVAILABLE FUNCTIONS (the only callables):
  SHOW_VARS()
  len(obj, /)
  llm_query(prompt, model=None)
  llm_query_batched(ps, model=None)
  rlm_query(prompt, model=None)
  rlm_query_batched(ps, model=None)
  str(...)
  print(...) — show values (output returned to you, truncated beyond 10000 chars)

CURRENT VARIABLES: answer, context
Variables persist across your code submissions, like notebook cells.

YOU CAN:
- assign and unpack variables:  parts = context.split(".")
- index and slice:              chunk = context[0:2000], last = items[-1]
- build literals and f-strings: q = f"summarize: {chunk}"
- use methods: all public methods on any object (e.g. .split() .strip() .count() .find()), except .format()/.format_map()
- use operators: string-shaped only: + (concat), == != in not in, and/or/not, ternaries (x if c else y). NO math (- * / // % < >) — delegate any computation
- echo a value by ending with a bare expression (like a REPL): parts[0]

YOU CANNOT (rejected before execution):
- loops, comprehensions, or generator expressions
- if statements (ternary expressions are OK)
- defining functions, lambdas, or classes
- imports, or calling anything not listed above
- attribute access outside method calls (no ._x, .__dunder__)
\end{promptbox}

\textbf{Full-Python run for Opus 4.8 on LongCoT-mini.} We report one RLM cell from the unrestricted variant, marked with an asterisk in Figure 2 and Table 1. Under the restricted workspace, Opus 4.8 on LongCoT-mini kept attempting to compute in the workspace instead of decomposing the problem, so we report its full-Python run. The other eight RLM cells use the restricted workspace.

\textbf{Budget awareness.} The reference implementation has no notion of a model-call budget. We add the same accounting the other agents use, charging every call at any depth against one allowance and injecting the same messages at the same thresholds. A system turn is appended each time another tenth of the allowance is consumed.

\begin{promptbox}
BUDGET: {used}/{budget} llmcall used, {remaining} remaining.
\end{promptbox}

At ninety percent of the allowance, one further message requires the model to submit.

\begin{promptbox}
⚠ FINISH NOW: budget is nearly exhausted. On this turn you must submit
your best answer by setting answer["content"] and answer["ready"] = True.
If you do not, the system will coerce a final answer, which may not be
your preferred one.
\end{promptbox}

The only difference from the messages the other agents receive is that the coercion names the \texttt{answer} dict, since RLM submits that way rather than by calling \texttt{finish}.

\section{Implementation details: Claude Code and Codex}
\label{app:coding}
\addcontentsline{atoc}{section}{\thesection\quad Implementation details: Claude Code and Codex}

Both coding agents are run in headless mode, \texttt{claude -p} and \texttt{codex exec}, and access the same per-problem container instance the Meta-Reasoning Agent and Direct Control Agent workers use through a single MCP tool, \texttt{container\_bash}. That tool reuses the same bash handler a worker gets, so both agents see identical output truncation, memory limits, and recovery behavior.

Their native tools are disabled so that every action flows through that one channel. Claude Code runs with its built-in tools off and only the MCP tool allowed; Codex runs with \texttt{--sandbox read-only}, so its own shell cannot write anything scored, with the MCP tool approved per-call. File edits therefore happen through heredocs and shell commands, exactly as they do for the Meta-Reasoning Agent and Direct Control Agent.

As with the other agents and systems, we keep these agents aware of the model-call budget. Each time another tenth of the allowance is consumed, a budget line is prefixed to the next tool result.

\begin{promptbox}
BUDGET: {used}/{budget} llmcall used, {remaining} remaining.
\end{promptbox}

The system prompt also states the allowance and that every turn costs one call.

\end{document}